\documentclass[10pt,twocolumn,letterpaper]{article}

\usepackage{wacv}
\usepackage{times}
\usepackage{epsfig}
\usepackage{graphicx}
\usepackage{amsmath}
\usepackage{amssymb}
\usepackage{booktabs}

\usepackage{comment}
\usepackage{subfigure}
\usepackage[export]{adjustbox}  
\usepackage[accsupp]{axessibility}  
\makeatletter
\@namedef{ver@everyshi.sty}{}
\makeatother

\usepackage{tikz}
\usetikzlibrary{spy,calc}
\newcommand{\method}[1]{\mbox{\textsc{#1}}}
\newcommand{\dataset}[1]{\mbox{\textsc{#1}}}
\newcommand{\mat}[1]{{\mathbf{#1}}}
\newcommand{\vect}[1]{{\mathbf{#1}}}
\newcommand{\T}{\intercal}

\newcommand*\circled[1]{\tikz[baseline=(char.base)]{
            \node[shape=circle,draw,inner sep=0.8pt] (char) {#1};}}

\graphicspath{{figures/}}

\def\wacvPaperID{385} 

\wacvalgorithmstrack   

\wacvfinalcopy 

\ifwacvfinal
\usepackage[breaklinks=true,bookmarks=false]{hyperref}
\else
\usepackage[pagebackref=true,breaklinks=true,colorlinks,bookmarks=false]{hyperref}
\fi

\begin{document}

\title{Planar Object Tracking via Weighted Optical Flow}

\author{Jon\'{a}\v{s} \v{S}er\'{y}ch, Ji\v{r}\'{i} Matas\\
  CMP Visual Recognition Group, Department of Cybernetics, \\
  Faculty of Electrical Engineering, Czech Technical University in Prague \\
{\tt\small \{serycjon,matas\}@fel.cvut.cz}
}

\maketitle
\thispagestyle{empty}

\begin{abstract}
  We propose WOFT -- a novel method for planar object tracking that estimates a full 8 degrees-of-freedom pose, \ie the homography w.r.t. a reference view.
  The method uses a novel module that leverages dense optical flow and assigns a weight to each optical flow correspondence, estimating a homography by weighted least squares in a fully differentiable manner.
  The trained module assigns zero weights to incorrect correspondences (outliers) in most cases, making the method robust and eliminating the need of the typically used non-differentiable robust estimators like RANSAC.
  The proposed weighted optical flow tracker (WOFT) achieves state-of-the-art performance on two benchmarks, POT-210~\cite{liang2017planar} and POIC~\cite{lin2019robust}, tracking consistently well across a wide range of scenarios.
\end{abstract}

\section{Introduction}
In this paper, we address the rigid planar object tracking problem, which is a specific subtopic of visual object tracking.
Given an object identified in the first frame, a tracker should output the tracked object pose or absence of the object in the frame, on every subsequent frame of a video sequence.
In a general model-free setting, the tracker has no prior knowledge about the target class except for target-specific information coming from the first frame initialization.
In standard tracking benchmarks, such as \dataset{OTB}~\cite{wu15otb}, \dataset{VOT}~\cite{kristan2018sixth,kristan2019seventh}, \dataset{LaSOT}~\cite{fan2019lasot}, \dataset{TrackingNet}~\cite{muller2018trackingnet}, or \dataset{YT-BB}~\cite{real2017youtube}, the object pose is represented by axis-aligned or rotated bounding boxes.
Tracking with segmentation mask representation has gained popularity in recent years, with benchmarks such as \dataset{VOT2020}~\cite{kristan2020eighth}, \dataset{DAVIS}~\cite{Perazzi2016,pont-tuset17davis}, and \dataset{YouTube-VOS}~\cite{xu2018youtube}.
\begin{figure}
  \centering
  \includegraphics[width=1\linewidth]{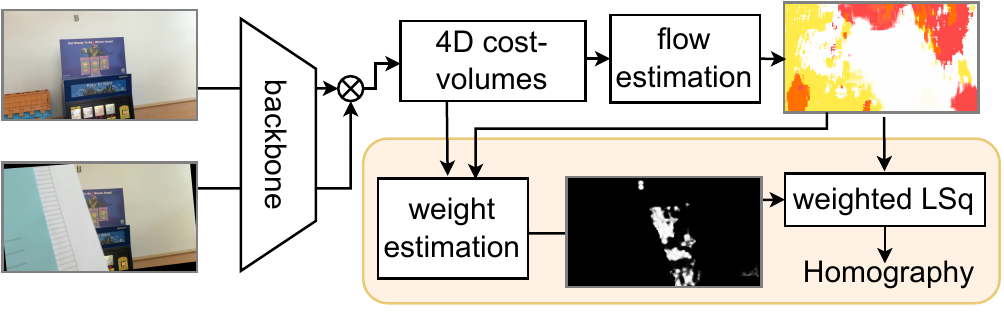}
  \caption{Planar object tracking with a homography estimated by a novel weighted least squares (LSq) homography module called \method{WFH} ({\it orange box}) on optical flow correspondences.
    The proposed trainable flow weight CNN assigns a weight $w_i \in [0, 1]$ to each flow vector based on samples from correlation cost-volume.
  }
  \label{fig:wraft}
\end{figure}

In planar rigid object tracking, the object pose is related to its initial pose by an 8 degrees-of-freedom (DoF) homography when using a perspective camera, and the target is fully specified by the initialization mask.
Planar trackers can output precise 8-DoF object poses, enabling applications not possible with bounding-box or segmentation mask level trackers, in areas such as film post-production, visual servoing~\cite{zhang2017asymptotic,benhimane2007homography}, SLAM~\cite{sun2019planar}, or markerless augmented reality~\cite{valognes2019augmenting,pirchheim2011homography,simon2000markerless}.
Man-made objects are often either completely planar or consist of planar surfaces, allowing for planar object tracking in a wide range of scenarios.

Current state-of-the-art methods struggle on seemingly toy-like sequences in standard planar object tracking datasets, \dataset{POT-210}~\cite{liang2017planar} and \dataset{POIC}~\cite{lin2019robust}.
The target planarity poses challenges, \eg, strong perspective distortion, significant illumination changes caused by specular highlights, and motion blur caused by a shaking hand-held camera.

In this work, we introduce a novel model-free planar object tracker.
The proposed method estimates dense correspondences between the template (initial image) and the current image with a deep optical flow network.
A novel homography estimation module then assigns a weight to each optical flow correspondence, and a homography is estimated as a solution to a weighted least squares problem.
The network assigns low weights to incorrect flow vectors and thus it is not necessary to use robust outlier detection algorithms like \method{RANSAC}.

Using dense OF correspondences has several advantages.
First, OF estimation is well researched and high-quality methods are available off-the-shelf.
Second, the dense per-pixel correspondences help on low-textured objects, where sparse key-point correspondences fail.
Finally, having dense correspondences enables us to compute a homography correspondence support set and detect a tracking failure if the support is small.

The proposed homography estimation procedure is fully differentiable, allowing us to train both the weight estimator and the optical flow network using homography supervision.
The main contributions of this work are the following.
\begin{itemize}
\item We propose a novel fully differentiable homography estimation neural network module.
\item We propose a novel planar target tracker employing the weighted flow homography estimation (code public\footnote{\url{https://cmp.felk.cvut.cz/~serycjon/WOFT}}).
\item The proposed tracker sets a new state-of-the-art on the \dataset{POT-210}~\cite{liang2017planar}, \dataset{POT-280}~\cite{liang2021planar}, and \dataset{POIC}~\cite{liang2017planar} datasets, performing well across all challenge types.
  On \dataset{POT-210}, the tracker error is half of the error of the best competing method.
\item We analyze the ground truth on the \dataset{POT-210} dataset and publish\addtocounter{footnote}{-1}\addtocounter{Hfootnote}{-1}\footnotemark{} a precise re-annotation of its subset.
  The inaccuracy of the original annotation accounted for half of the errors of the proposed tracker.
\end{itemize}

\section{Related Work}
General visual object tracking methods have been improving consistently, with deep-learning-based trackers dominating classical methods\cite{kristan2021vot,kristan2020eighth}.
In contrast, planar object trackers have only recently started using deep learning.

The homography trackers can be roughly divided into three main categories~\cite{liang2017planar}: keypoint methods, direct methods, and deep methods.
Traditional keypoint-based tracking consists of three steps: (i) keypoint detection and description using, \eg, \method{SIFT}\cite{lowe2004distinctive} or \method{SURF}\cite{bay2008speeded}, (ii) keypoint matching by nearest neighbor search in the descriptor space, and (iii) robust homography estimation with RANSAC~\cite{fischler1981random}.
The \method{SOL}~\cite{hare2012efficient} tracker uses SVM to learn keypoint descriptors and PROSAC~\cite{chum2005matching} ordering.
In \method{Gracker}~\cite{wang2017gracker}, the keypoints are not matched independently based only on descriptor similarity, but instead a graph-matching approach is used.
The \method{OBD}~\cite{matveichev2021mobile} tracker uses ORB keypoints for target detection and optical flow tracking.
In the \dataset{POT-280}~\cite{liang2021planar} benchmark, the authors compare several deep-learning based homography trackers.
The best ones use the \method{SIFT} keypoint detector, a deep learning descriptor such as \method{GIFT}~\cite{liu2019gift}, \method{MatchNet}~\cite{han2015matchnet}, \method{SOSNet}~\cite{tian2019sosnet}, or \method{LISRD}~\cite{pautrat2020online}, followed by \method{RANSAC}.

Direct methods formulate the tracking task as image registration.
Given the current frame, they attempt to find a homography warping that optimizes the alignment of the current frame with the object in the initial frame.
In the classical Lucas-Kanade~\cite{lucas1981iterative} and the Inverse Compositional~\cite{baker2004lucas} methods, the warp quality is measured directly on the image intensities by a sum of squared differences.
The \method{ESM}~\cite{benhimane2004real} tracker avoids the costly computation of Hessian in Lucas-Kanade by using an efficient second-order minimization (ESM) technique.
\method{GO-ESM}~\cite{chen2017illumination} improved robustness to illumination changes by adding a gradient orientation feature on top of the image intensities and generalizing the \method{ESM} tracker to multidimensional features.
The \method{GOP-ESM}~\cite{lin2019robust} tracker extends \method{GO-ESM} with a feature pyramid and a coarse-to-fine iterative approach.
Chen~\etal~\cite{chen2019learning} proposed to use the ESM algorithm as a differentiable layer in a siamese neural network architecture.
The ESM layer iteratively aligns the template and the query frame feature maps obtained from a \method{ResNet-18}~\cite{he15resnet} backbone pre-trained on \dataset{ImageNet}.
The whole architecture is then fine-tuned on image pairs synthesized from the MS-COCO dataset~\cite{lin14coco}. 
Direct methods perform well on the \dataset{POIC}\cite{lin2019robust} dataset, but typically fail on motion blur, partial occlusions and partially out-of-view targets, \eg in the \dataset{POT-210}~\cite{liang2017planar} dataset.

Deep learning homography estimation is typically done by regression of four control points.
The \method{HomographyNet}~\cite{detone2016deep} and \method{UDH}~\cite{nguyen2018unsupervised} feed a concatenated pair of homography-related images through a CNN and formulate the homography estimation as direct regression
of four control points.
Rocco~\etal~\cite{rocco2017convolutional} proposed to regress the four homography control points from a correlation cost-volume containing each-to-each similarities between Siamese VGG-16~\cite{simonyan14vgg} feature maps.
The four-point regression is also used by the recently proposed \method{HDN}~\cite{zhan2021homography} method, which decomposes the homography into a similarity transform and a homography residual.
These control-point regression methods struggle with occlusions and often assume that the whole images are related by a homography, and do not distinguish between the target and the background motion.
The \method{PFNet}~\cite{zeng2018rethinking} uses a custom convolutional architecture to estimate a dense flow field, which is then used in \method{RANSAC}, making the method not differentiable and end-to-end training not possible.

\section{Method}
We propose a weighted flow homography module (\method{WFH}) that assigns a flow weight $w_i \in [0, 1]$ to each OF correspondence and estimates a homography using a weighted least squares formulation (Sec.~\ref{sec:WFH}).
The \method{WFH} is differentiable, making end-to-end training of both the \method{WFH} and the OF network possible.
In Sec.~\ref{sec:homography-tracker}, we propose a weighted optical flow tracker (\method{WOFT}) built around the \method{WFH} homography estimator.

\subsection{Weighted Flow Homography Module}
\label{sec:WFH}
The idea of the \method{WFH} module is to compute a \emph{flow weight} $w_{i} \in [0, 1]$ for each optical flow vector and to predict a homography by solving a weighted least squares (LSq) problem.
The standard least squares homography fitting is sensitive to grossly incorrect correspondences (outliers).
This is usually addressed by \method{RANSAC}, which uses repeated hypothesis sampling to find a homography and its outlier-free correspondence support set.
The \method{WFH} instead eliminates outliers by setting their flow weights close to zero, allowing for a robust, single iteration, and fully differentiable weighted least squares fitting.

We process a pair of images with an optical flow estimation network, such as \method{RAFT}~\cite{teed2020raft} to get OF correspondences $\left( \vect{p}_i, \vect{p^{\prime}}_i \right)$,
where $\vect{p}_i = ( x_i, y_i )$ is a position in one image and $\vect{p}_i^{\prime} = (x_i^{\prime}, y_i^{\prime})$ the corresponding position in the second image.
We then pass a suitable inner representation of the OF network to a weight-estimation CNN that predicts the flow weight $w_i$ for each OF vector.
Finally, we estimate homography by solving a system of equations by weighted least squares.
First, we introduce plain least squares homography estimation, then we describe the weighted variant and the training loss function.
Finally, we describe the weight estimation CNN in detail.
\\ \textbf{LSq Homography}.  Given the optical flow correspondences, we want to find a homography matrix $\mat{H} \in \mathbb{R}^{3 \times 3}$ mapping $\left( x_i, y_i, 1 \right)$ to $\left( \lambda x_i^{\prime}, \lambda y_i^{\prime}, \lambda \right)$, $\lambda \neq 0$.
This leads to an overdetermined homogeneous system of equations $\mat{A} \vect{h} = \vect{0}$, with $\vect{h} \in \mathbb{R}^{9 \times 1}$ being the flattened $\mat{H}$-matrix and $\mat{A} \in \mathbb{R}^{2N \times 9}$ encoding the data constraints.
The system is usually solved in the least-norm sense via a singular value decomposition (SVD) of $\mat{A}$.
We use the PyTorch machine learning framework which includes differentiable SVD, but the back-propagated gradients are often unstable.
To overcome this issue, we constrain the homography by fixing its bottom-right element $h_{3,3} = 1$, leading to a non-homogeneous system $\tilde{\mat{A}} \tilde{\vect{h}} = \vect{b}$, which can be solved in the least-squares sense using the QR decomposition with more stable gradients.
Not all homographies are representable with this constraint (see Sec. 4.1.2 in \cite{hartley04thebook}), but we have not encountered such a case in the tracking scenario.

In the non-homogeneous formulation, each correspondence adds two equations into $\tilde{\mat{A}} \in \mathbb{R}^{2N \times 8}$ and $\vect{b} \in \mathbb{R}^{2N}$:
\begin{equation}
  \label{eq:non-homogeneous-system}
  \begin{bmatrix}
    0 & 0 & 0 & -x_i & -y_i & -1 & y_i^{\prime} x_i & y_i^{\prime} y_i \\
    x_i & y_i & 1 & 0 & 0 & 0 & -x_i^{\prime} x_i & -x_i^{\prime} y_i \\
  \end{bmatrix} \tilde{\vect{h}} =
  \begin{bmatrix}
    -y_i^{\prime} \\
    x_i^{\prime}
  \end{bmatrix}
\end{equation}
\begin{figure}
  \centering
  \includegraphics[trim=150 150 200 90,clip,width=0.49\linewidth]{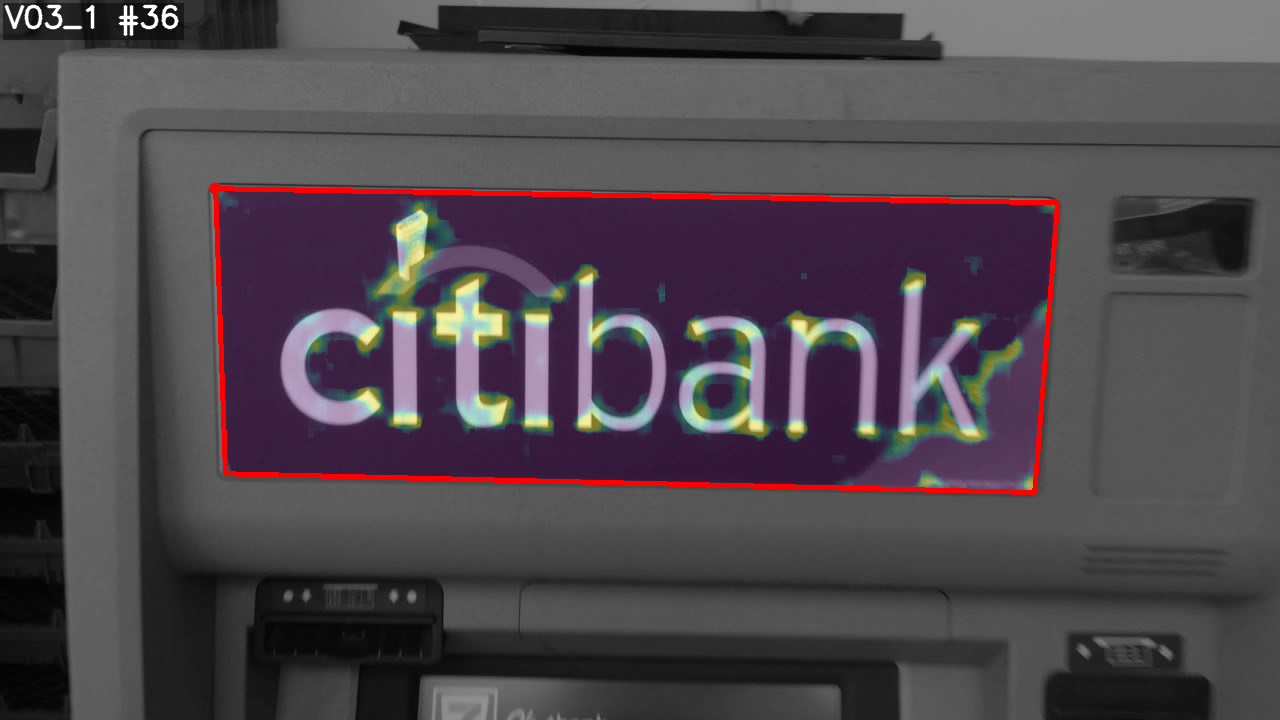}
  \includegraphics[trim=200 140 200 120,clip,width=0.49\linewidth]{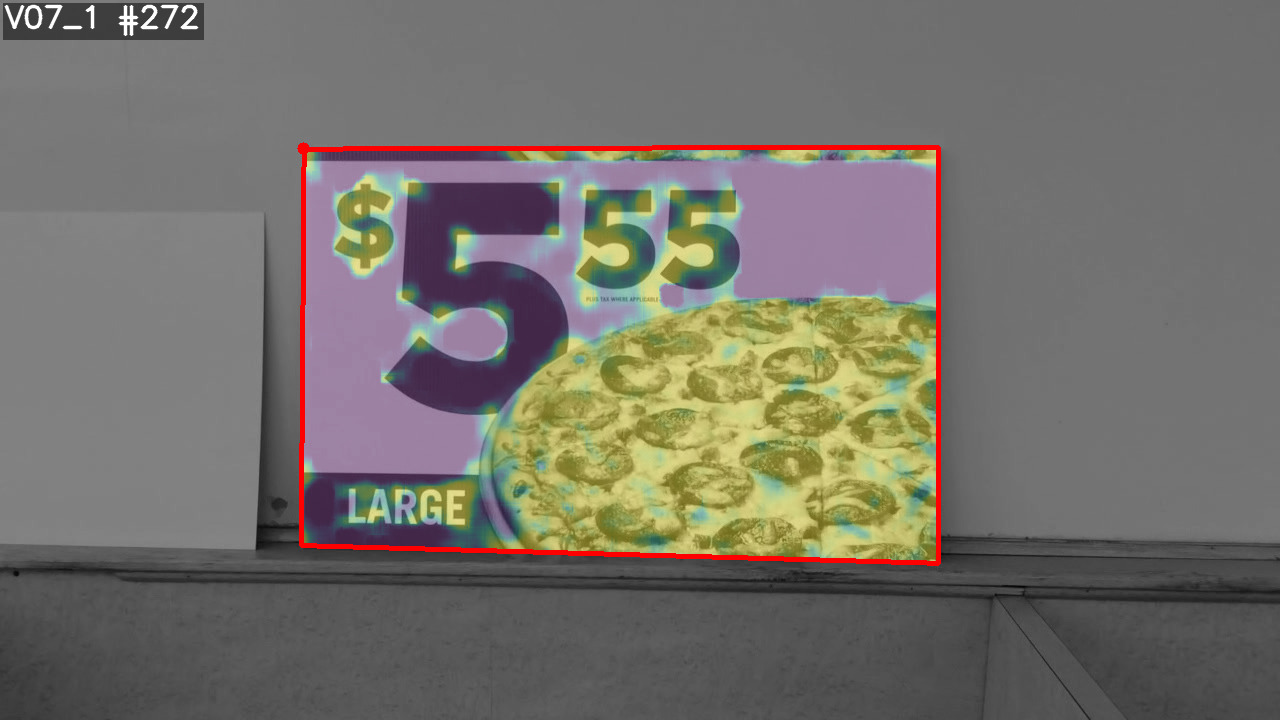}
  \includegraphics[trim=300 50 200 40,clip,width=0.49\linewidth]{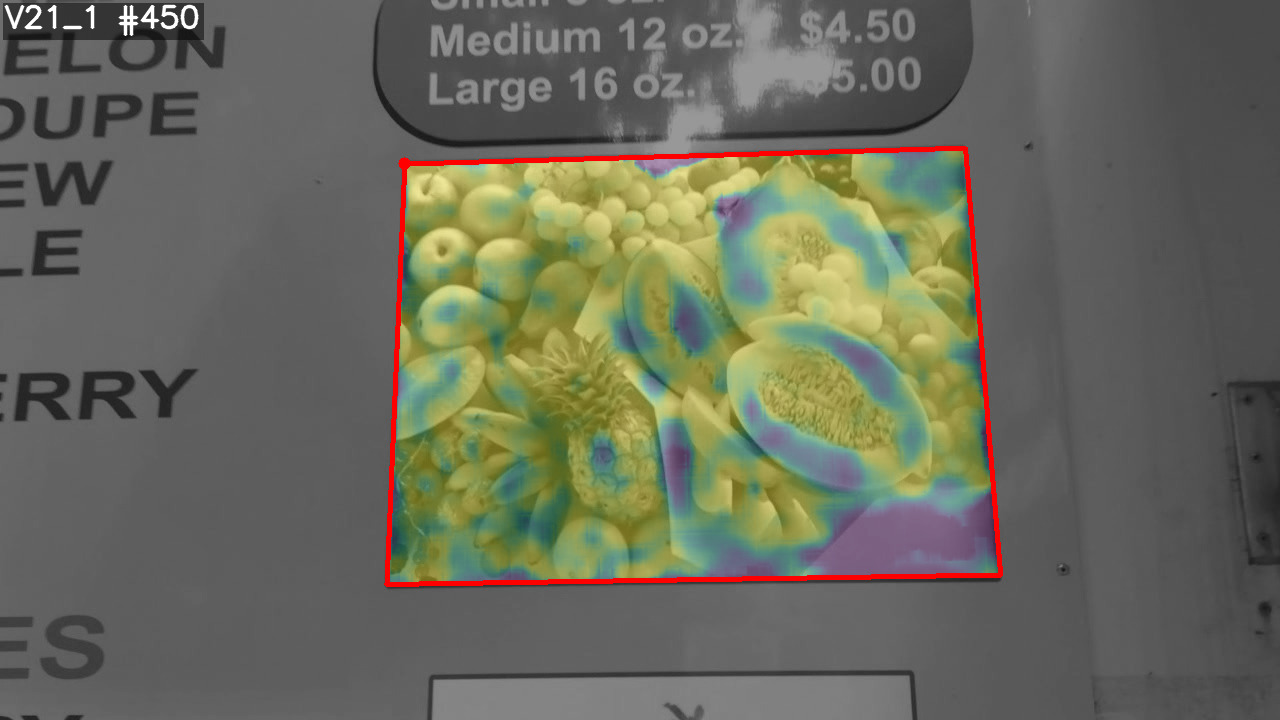}
  \includegraphics[trim=350 140 400 150,clip,width=0.49\linewidth]{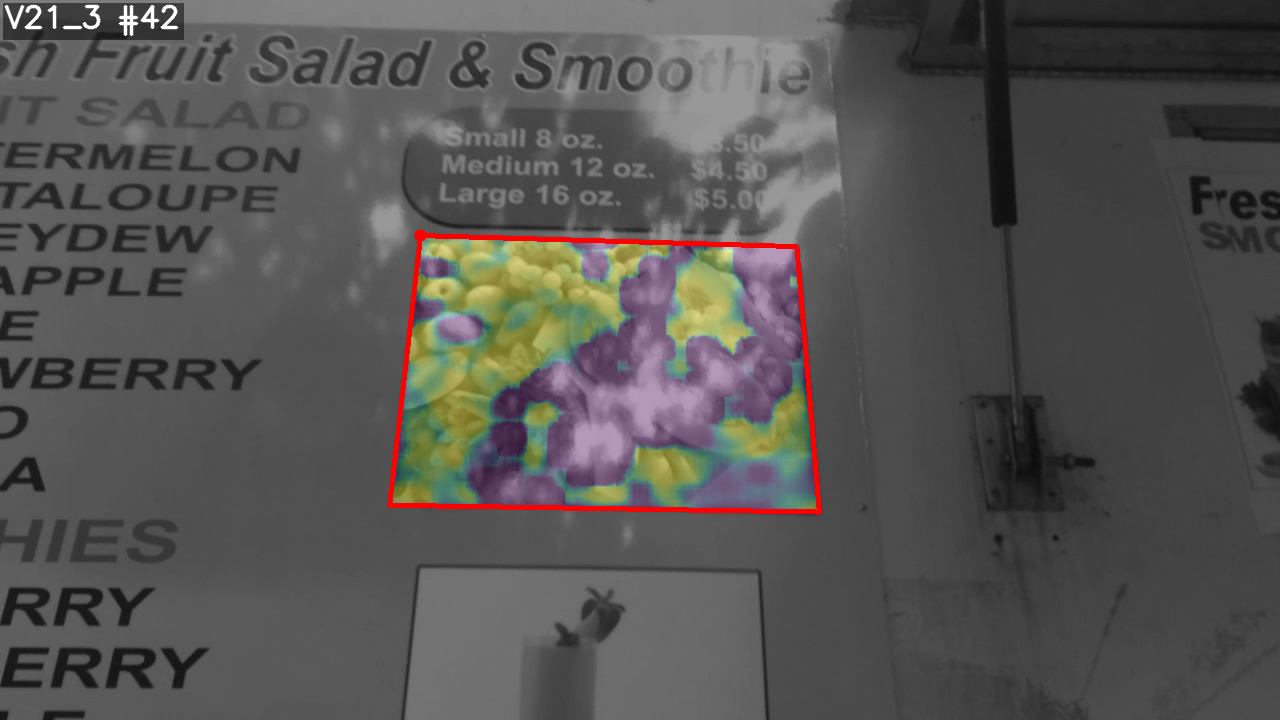}
  \caption{High weights of optic flow ({\it yellow}) appear mainly on corners and well-textured areas.
    {\it Bottom}: the \dataset{POT-210} target with the highest average flow weights ({\it left}); weight values drop ({\it purple}) when ''occluded'' by a reflection  ({\it right}).
    Best viewed in color.}
  \label{fig:example-weights}
\end{figure}
We solve the least squares problem
\begin{equation}
  \label{eq:least-squares}
\min\limits_{\tilde{\vect{h}}} \sum\limits_{j=1}^{2N} \lVert \tilde{\mat{A}}_{j,\cdot} \tilde{\vect{h}} - \vect{b}_j \rVert_2^2
\end{equation}
by QR decomposition of the data matrix $\tilde{\mat{A}} = \mat{Q} \mat{R}$ followed by solving the triangular system $\mat{R} \tilde{\vect{h}} = \mat{Q}^{\T} \vect{b}$ (triangular system solver available in PyTorch).
\\ \textbf{Weighted LSq Homography}.  In the proposed weighted least squares formulation, we weight each pair of equations with the corresponding estimated flow weight $w_i$ and find
\begin{eqnarray}
  \label{eq:weighted-least-squares}
  \min\limits_{\tilde{\vect{h}}} \sum\limits_{j=1}^{2N} w_j \lVert \tilde{\mat{A}}_{j,\cdot} \tilde{\vect{h}} - \vect{b}_j \rVert_2^2\\
   = \min\limits_{\tilde{\vect{h}}} \sum\limits_{j=1}^{2N} \lVert \left( \sqrt{w_j} \tilde{\mat{A}} \right)_{j,\cdot} \tilde{\vect{h}} - \left( \sqrt{w_j}\vect{b}_j \right) \rVert_2^2 
\end{eqnarray}
The weighted problem~\eqref{eq:weighted-least-squares} is transformed into non-weighted (Eq.~\eqref{eq:least-squares}) by multiplying each row of $\tilde{\mat{A}}$ and each element of $\vect{b}$ by the square root of the corresponding weight~$\sqrt{w_i}$.
\\ \textbf{Training WFH.}  We train the \method{WFH} weight estimation CNN using a loss function on the predicted homography.
We warp points forward by the ground truth homography $\mat{H}_{GT}$ then backward by the inverse of the estimated $\mat{H}$ and finally compute L1 loss on the projection errors as:
\begin{equation}
  \label{eq:loss}
  L(\mat{H}) = \frac{1}{N} \sum\limits_{i=1}^N \lVert  p_i - \mat{H}^{-1}\mat{H}_{GT} p_i \rVert_2
\end{equation}

Both the optical flow network and the flow weight estimation CNN are trained using a single loss function, and we do not use additional direct supervision of the flow weight predictor.
The learned flow weights resemble a keypoint detector output (corners, well-textured patches), but with information from both images, therefore giving low weights on occlusions or significant appearance changes as shown in figure~\ref{fig:example-weights}.
\\ \textbf{Weight Estimation CNN}
The proposed \method{WFH} module operates on the correlation cost-volume pyramid of the \method{RAFT}~\cite{teed2020raft} optical flow estimator, but the idea is applicable to other OF networks (Sec.~\ref{sec:ablation-study}).
\method{RAFT} computes a correlation volume $\mathbf{C}^1 \in \mathbb{R}^{H/8 \times W/8 \times H/8 \times W/8}$ that captures the similarity between all pairs of feature vectors extracted from the two input images.
Next, they construct a 4-layer correlation pyramid $\left\{ \mathbf{C}^1,\mathbf{C}^2,\mathbf{C}^3,\mathbf{C}^4 \right\}$.
Finally, $9 \times 9$ patches centered on current flow vector estimates are sampled from this pyramid and processed by a neural network that produces a flow vector update.
This is repeated several times to produce the final optical flow field.

In \method{WFH} we sample the correlation pyramid once more on the final OF positions, resulting in a $9 \times 9 \times 4$ feature map for each flow vector in the spatial resolution of $H/8 \times W/8$.
To capture the global context, we then append an additional channel containing the mean correlation volume response $M(i, j) = \sum\limits_{m=1}^{H/8}\sum\limits_{n=1}^{W/8} \mathbf{C}_{i, j, k, l}^1$ for the given position $(i, j) \in \{1, \dots, H/8\} \times \{1, \dots, W/8\}$ in the first input image feature map.
We process the resulting features $f_{i, j} \in \mathbb{R}^{9 \times 9 \times 5}$ with a three-layer convolutional network (kernel size 3, 128 output channels, ReLU) followed by a $1 \times 1$ convolution (single output channel) and global average pooling.
Finally, we up-sample the results with the \method{RAFT} up-sampling module and apply a sigmoid activation to get a $H \times W$ score map with weights between $0$ and $1$.

\subsection{Homography tracker}
\label{sec:homography-tracker}
\begin{figure}
  \centering
  \includegraphics[width=1\linewidth]{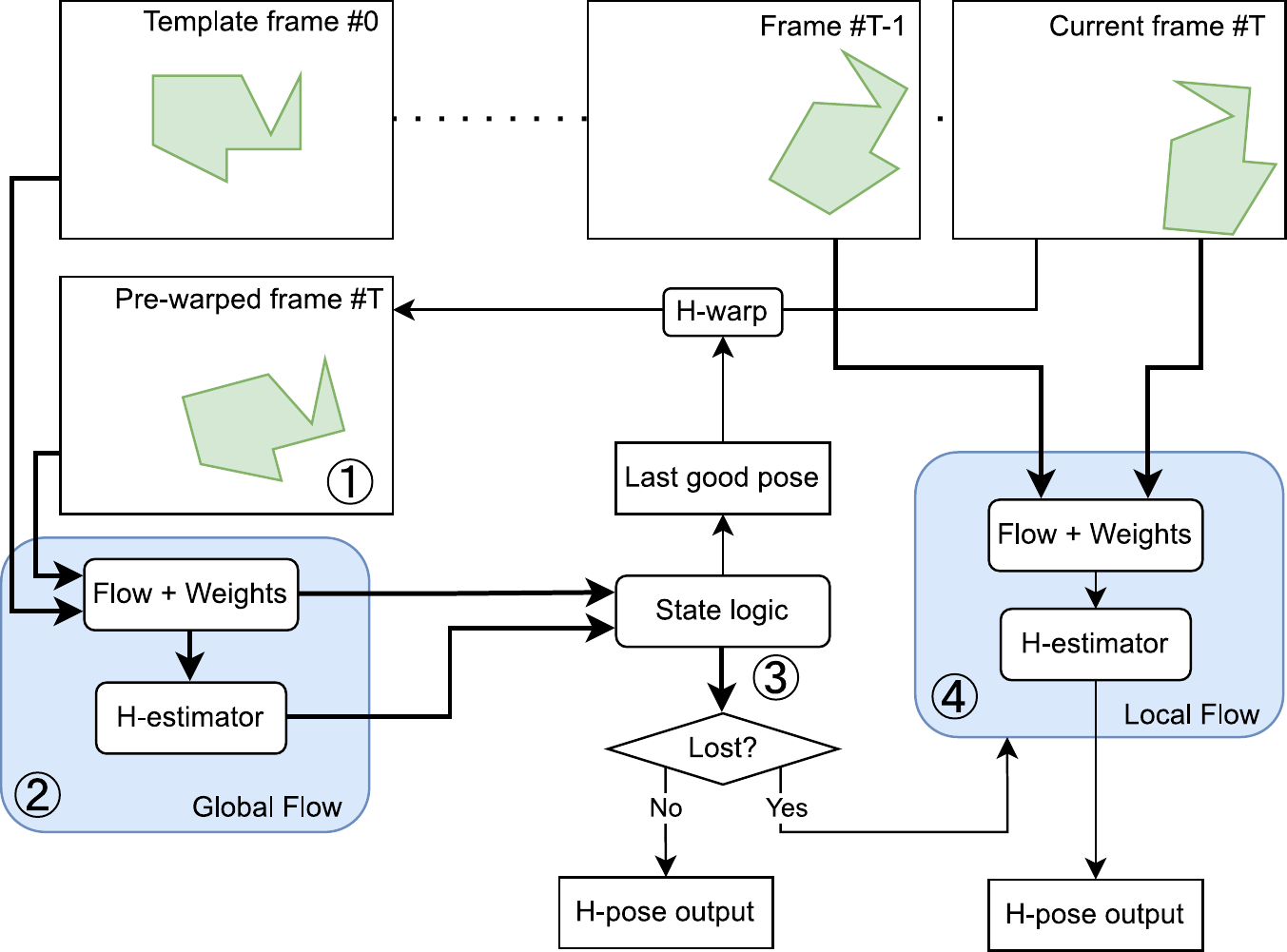}
  \caption{The \method{WOFT} tracker pre-warps the current frame using the last good pose (1).
    It then estimates a homography between the template and the pre-warped frame (2) and
    the reliability of the estimated homography is assessed (3).
    When the estimate is not reliable (`lost` state) a homography based on a local flow (4) is returned instead.
  }
  \label{fig:tracker-overview}
\end{figure}
We propose a planar object tracker based on the weighted flow homography module, \method{WFH}.
Our weighted optical flow tracker, denoted \method{WOFT}, consists of four main parts as shown in~Fig.~\ref{fig:tracker-overview}.

First, we apply a pre-warping technique to reduce large pose differences, which are not handled well by OF methods.
The current video frame $I_t$ is pre-warped \circled{1} by the homography from the last reliable frame $I_G$, with $G = 0$ initially.
The pre-warp $\tilde{I}_t = \mathcal{W}(\mat{H}_{0 \rightarrow G}^{-1}, I_t)$ \circled{1} reduces the pose difference between the template and the current images, resulting in a motion similar to the typical $I_{(t-1)} \rightarrow I_t$ optical flow scenario.
The possible appearance difference between the template and a temporarily distant frame (caused mainly by illumination changes and motion blur) is implicitly handled by the optical flow feature encoder.

Second, we compute the \emph{global} optical flow \circled{2} between the template frame $I_0$ and the pre-warped current frame $\tilde{I}_{t}$ and the corresponding flow weights.
We mask the flow correspondences, only leaving the ones starting inside the template mask and ending inside the current image.
To speed up the homography estimation, we randomly subsample the correspondences, only keeping 500.
We then estimate homography $\mat{H}_{0 \rightarrow \tilde{t}}$ using weighted least squares as described in Sec.~\ref{sec:WFH}.
Computing the homography between the template and the pre-warped current frame prevents error accumulation and target drift (Sec.~\ref{sec:ablation-study}).

We pass the weighted optical flow together with the computed homography to a state logic block \circled{3} that decides whether the tracking was successful or not.
The lost/not-lost decision is made based on the support set size of the estimated homography.
In particular, with optical flow correspondences $(\vect{p}_i, \vect{p}_i^{\prime})$ we warp each position $\vect{p}_i = (x_i, y_i)$ using the homography $\mat{H}_{0 \rightarrow \tilde{t}}$ and compute the Euclidean distance to the position $\vect{p}_i^{\prime} = (x_i^{\prime}, y_i^{\prime})$.
The $i$-th correspondence is an inlier when $\lVert \mathcal{W}(\mat{H}_{0 \rightarrow \tilde{t}}, \vect{p}_i) - \vect{p}_i^{\prime} \rVert \leq 5$ pixels -- a standard threshold on planar tracking benchmarks~\cite{liang2017planar,lin2019robust}.
We declare the tracker lost when it has a small support set, \ie less than $20\%$ inliers.

When the tracker is not lost, we return $\mat{H}_{0 \rightarrow t} = \mat{H}_{0 \rightarrow G}^{-1} \mat{H}_{0 \rightarrow \tilde{t}}$ and update the last good frame index used for pre-warping $G = t$.
When the tracker is lost, we make a second attempt to estimate the pose using a \emph{local} optical flow $I_{(t-1)} \rightarrow I_t$.
The local flow tends to drift, but it helps to keep track of the target pose in the short term.
The temporarily close input images are close in appearance (similar illumination, similar motion blur, \etc).
We estimate $\mat{H}_{(t-1) \rightarrow t}$ \circled{4} by weighted least squares as described above
and output $\mat{H}_{0 \rightarrow t} = \mat{H}_{(t-1) \rightarrow t} \mat{H}_{0 \rightarrow (t-1)}$.
Moreover, when the tracker is lost for more than 10 frames, we reset the pre-warping last good frame index $G = 0$.
The target pose can change significantly over the 10 frames, making the pre-warp information outdated.
Moreover, a bad pre-warp homography can ruin any chance of recovering, \eg an outdated strong perspective change pre-warp distorts the current target area beyond being recognizable, and the identity homography with $G = 0$ is the safest choice.

\subsection{Implementation details}
For optical flow, we use the author-provided \method{RAFT} checkpoint trained on Sintel.
We then train the weight estimation CNN for 10 epochs on a synthetic dataset with 50000 image pairs.
We generate the training set by repeatedly sampling a random \dataset{MS COCO}\cite{lin14coco} image and warping it with two random homographies representing the template and the current frame pose.
The random homographies are generated by perturbing each corner of the image with a random vector of length up to $20\%$ of the image diagonal.
We blur the second warped image by a random linear motion of length up to 20 pixels.
Finally, both images are passed through JPEG compression with quality set to 25.

We train with \method{AdamW}~\cite{loshchilov2018decoupled} optimizer with an initial learning rate $1e^{-3}$, which is then halved after every epoch.
Finally, we fine-tune the whole network, including \method{RAFT} for 2 epochs, starting from the learning rate $1e^{-5}$ and again halving it after every epoch.
To stabilize the training procedure, we discard training samples achieving loss over 100.

The tracker runs at around 3.5 FPS on a GeForce RTX 2080 Ti GPU (i7-8700K CPU @ 3.70GHz).
The majority of time is spent on the optical flow computation (275ms).
The weight computation (2ms), the weight up-sampling (1ms), and the least squares homography estimation (5ms) take negligible time.
Image pre-warping (done on CPU), optical flow masking, and subsampling cost an additional 7ms.

A faster variant \method{WOFT}\textsubscript{$\downarrow\! s$} downscales the input images to $H/s \times W/s$ and rescales the output homographies to the original resolution.

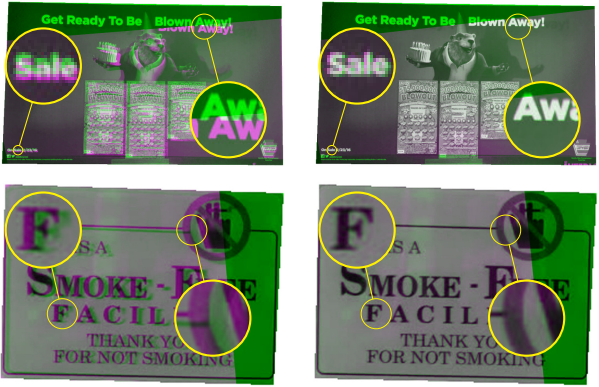
\begin{figure}
  \newlength{\potwidth}
  \setlength{\potwidth}{0.49\linewidth}

  \centering
  \begin{tikzpicture}[spy using outlines={circle,yellow,magnification=2.5,size=1cm, connect spies}]
    \node[anchor=south west,inner sep=0] (img)  at (0,0) {\includegraphics[width=\potwidth]{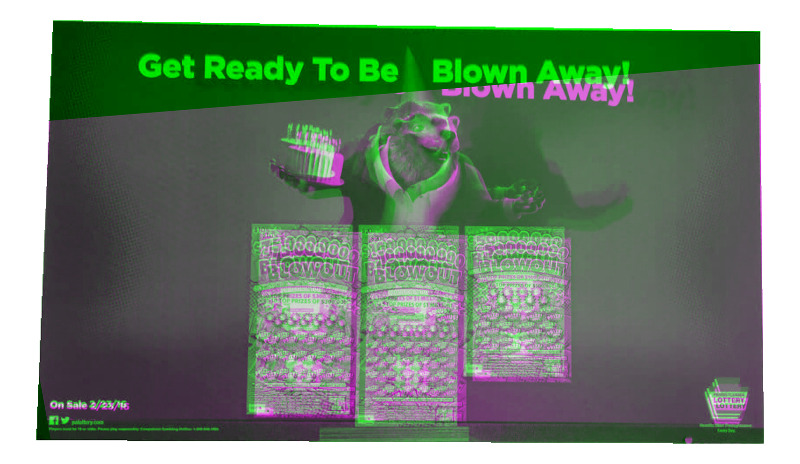}};
    \begin{scope}[x={($ (img.south east) - (img.south west) $ )},y={( $ (img.north west) - (img.south west)$ )}, shift={(img.south west)}]
      \node (spy1) at (0.096, 0.13) {};
      \node (spy1to) at (0.3, 0.6) {};
      \spy[magnification=8] on (spy1) in node [left] at (spy1to);
      \node (spy2) at (0.70, 0.82) {};
      \node (spy2to) at (0.90, 0.3) {};
      \spy[magnification=3] on (spy2) in node [left] at (spy2to);
    \end{scope}
  \end{tikzpicture}
  \begin{tikzpicture}[spy using outlines={circle,yellow,magnification=5,size=1cm, connect spies}]
    \node[anchor=south west,inner sep=0] (img)  at (0,0) {\includegraphics[width=\potwidth]{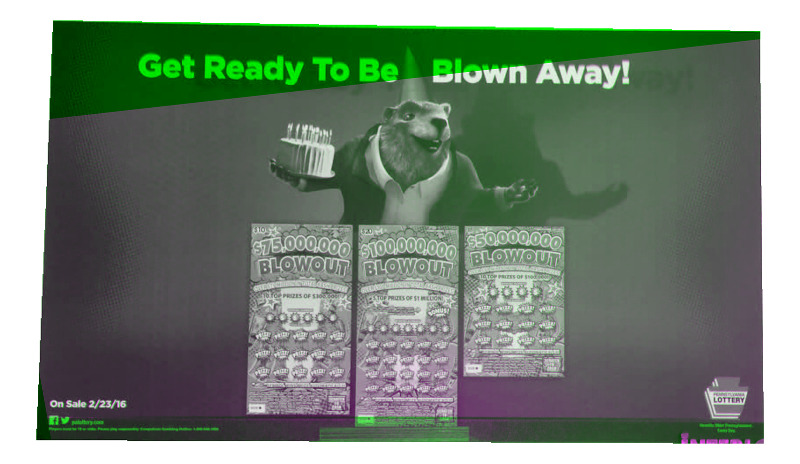}};
    \begin{scope}[x={($ (img.south east) - (img.south west) $ )},y={( $ (img.north west) - (img.south west)$ )}, shift={(img.south west)}]
      \node (spy1) at (0.096, 0.13) {};
      \node (spy1to) at (0.3, 0.6) {};
      \spy[magnification=8] on (spy1) in node [left] at (spy1to);
      \node (spy2) at (0.70, 0.82) {};
      \node (spy2to) at (0.90, 0.3) {};
      \spy[magnification=3] on (spy2) in node [left] at (spy2to);
    \end{scope}
  \end{tikzpicture}
  \begin{tikzpicture}[spy using outlines={circle,yellow,magnification=2.5,size=1cm, connect spies}]
    \node[anchor=south west,inner sep=0] (img)  at (0,0) {\includegraphics[width=\potwidth]{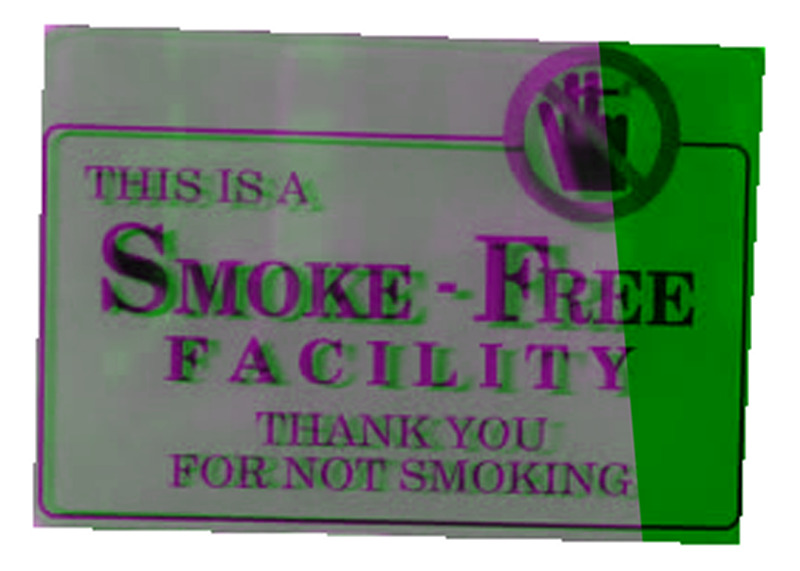}};
    \begin{scope}[x={($ (img.south east) - (img.south west) $ )},y={( $ (img.north west) - (img.south west)$ )}, shift={(img.south west)}]
      \node (spy1) at (0.24, 0.37) {};
      \node (spy1to) at (0.3, 0.75) {};
      \spy on (spy1) in node [left] at (spy1to);
      \node (spy2) at (0.66, 0.75) {};
      \node (spy2to) at (0.85, 0.35) {};
      \spy on (spy2) in node [left] at (spy2to);
    \end{scope}
  \end{tikzpicture}
  \begin{tikzpicture}[spy using outlines={circle,yellow,magnification=2.5,size=1cm, connect spies}]
    \node[anchor=south west,inner sep=0] (img)  at (0,0) {\includegraphics[width=\potwidth]{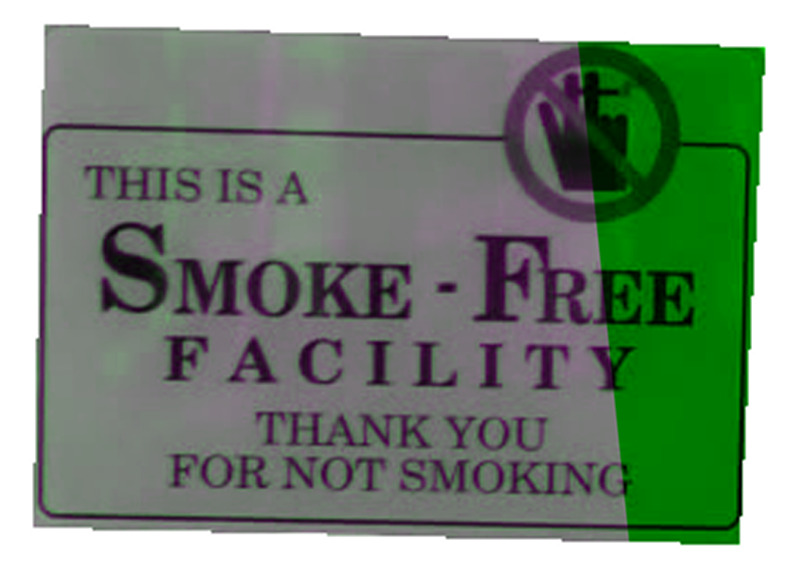}};
    \begin{scope}[x={($ (img.south east) - (img.south west) $ )},y={( $ (img.north west) - (img.south west)$ )}, shift={(img.south west)}]
      \node (spy1) at (0.24, 0.37) {};
      \node (spy1to) at (0.3, 0.75) {};
      \spy on (spy1) in node [left] at (spy1to);
      \node (spy2) at (0.66, 0.75) {};
      \node (spy2to) at (0.85, 0.35) {};
      \spy on (spy2) in node [left] at (spy2to);
    \end{scope}
  \end{tikzpicture}
  \caption{Precise re-annotation examples. 
  Original ground truth annotation ({\it left}), improved ground truth annotation ({\it right}).
    The grayscale template in green channel, the GT-warped current frame in red and blue channels.
    Imprecise annotation causes green and magenta shadows, while precisely aligned images produce a grayscale result.
  The green bands on top and on right side respectively are caused by a partial occlusion on current frame.
      The alignment error of the original GT evaluated on the improved ground truth is $15.8$px ({\it top}) and $7.2$px ({\it bottom}).
  }
  \label{fig:reannotation}
\end{figure}

\section{Experiments}
We evaluate the proposed tracker on two standard planar object tracking datasets, \dataset{POT-210} and \dataset{POIC} and show that it consistently achieves high accuracy and robustness.
\\ \textbf{POT-210\cite{liang2017planar}:} The Planar Object Tracking in the Wild benchmark contains 210 videos of 30 objects.
Each object appears in 7 video sequences with different challenging attributes -- \emph{scale} change, in-plane \emph{rotation}, \emph{perspective} distortion,
motion \emph{blur}, \emph{occlusion}, \emph{out-of-view}, and \emph{unconstrained}.
The sequences have a fixed length of 501 frames.
\dataset{POT-280}~\cite{liang2021planar} extends \dataset{POT-210} by 10 new objects.
\\ \textbf{POIC\cite{lin2019robust}:} the Planar Objects with Illumination Changes dataset consists of 20 sequences of varying length giving a total of 22971 frames.
The dataset contains sequences with translation, in- and out-of-plane rotations, and scale changes, but mainly focuses on strong specular highlights and other significant illumination changes, making it complementary to \dataset{POT-210}.
\\ \textbf{Evaluation protocol:}
On both \dataset{POT-210} and \dataset{POIC}, a tracker is initialized on the first frame and left to track till the end of the sequence.
The \emph{alignment error} $e_{AL}$ is computed for each annotated frame.
Given four reference points $\vect{x}_i \in X$ in the first frame, the alignment error is defined as root-mean-square error between their projection into the current frame by the ground truth homography $\mat{H}^*$ and by the tracker homography $\mat{H}$,
\begin{equation}
  \label{eq:AE}
  e_{AL}(\mat{H}; \mat{H}^{*}, X) = \sqrt{
    \frac{1}{4} \sum\limits_{i=1}^4
      \left(\mathcal{W}(\mat{H}^{*}, \vect{x}_i) - \mathcal{W}(\mat{H}, \vect{x}_i\right))^2},
\end{equation}
with $\mathcal{W}(\mat{H}, \vect{x})$ representing the projection of vector $\vect{x}$ by a homography $\mat{H}$.
Tracker precision is measured as a fraction of frames with $e_{AL} \leq 5$~px (P@5 score).
Additionally, we measure $e_{AL} \leq 15$~px (P@15 score), corresponding to the fraction of frames with target not tracked perfectly, but not completely lost either -- we call this robustness regime.

\subsection{Ground truth quality}
During the analysis of \method{WOFT} performance on \dataset{POT-210}, we found that in many cases the ground truth (GT) annotations are less accurate than the official 5px error threshold.
We have performed reannotation of a subset of the \dataset{POT-210} dataset to measure the original GT quality and provide more accurate estimates of tracker performance, see Fig.~\ref{fig:reannotation}.
Our annotation tool shows the template, the object on the current frame warped with the current annotation, and, most importantly, an alignment visualization.
We convert both the template frame and the current frame to grayscale and overlay the warped frame over the template, putting the template into the green channel and the current frame into the red and the blue channels.
This allows for very precise alignment over the whole extent of the target, unlike the annotation interface used for the original annotation (Fig.~4 in~\cite{liang2017planar}).
We have fully manually reannotated frames 82, 172, 252, 332, and 412 from each sequence, without seeing the \method{WOFT} estimated poses and the new GT will be made publicly available.
More examples of the reannotation overlay are in the supplementary materials.
The alignment error of the original GT evaluated on our re-annotation is $3.63$ on average, and worse than the official $5$px threshold in $15\%$ cases.

\subsection{Ablation study}
\label{sec:ablation-study}
In Table~\ref{tab:ablation}, we show the impact of various design choices of \method{WOFT} on \dataset{POT-210} performance (both on the original and the more accurate re-annotated ground truth).
First, we show the importance of computing the optical flow between the template and the pre-warped current frame.
In rows 1, 2 we only use the local flow (from $I_{(t-1)}$ to $I_t$).
The tracker drifts and quickly loses the target, resulting in overall poor performance.
A big performance improvement is achieved by using global flow (from $I_0$ to $\tilde{I}_t$) and always using the previous frame for pre-warping (rows 3, 4).
Another boost in performance is achieved with the controlled pre-warping (rows 5 - 9), where the local flow is used when the global flow fails and the pre-warp homography is reset when the target is `lost` for more than 10 frames. 

Using the weighted least squares homography estimation consistently improves the performance -- compare row 2 to row 1 (P@5 $+1.3$), row 4 to row 3 (P@5 $+10.7$), and row 6 to row 5 (P@5 $+8.3$).
In row 7, we used the same settings as in \method{WOFT} (row 6), but without the \method{RAFT} fine-tuning, resulting in a drop in P@5 ($-7.4$).
We have also experimented (row 8) with estimating homography by weighted iterative reweighted least squares (IRLSq) instead of ordinary weighted least squares.
We have set the IRLSq to optimize the Huber loss (also called smooth L1 loss) which is more robust to outliers than least squares.
This did not change the performance (w.r.t.~row 6), indicating that our estimated weights already take care of outliers and the robust estimator is not necessary.
Next, we compare \method{RANSAC} (row 9) with the proposed \method{WOFT} (row 6).
The weighted least squares approach achieves better results (P@5 $+0.9$) in a single differentiable pass.

Rows 10-12 show \method{WOFT} with \method{LiteFlowNet2}~\cite{hui2020lightweight} flow (details in supplementary) instead of \method{RAFT}.
Again, the weighted LSq estimator (row 12) works better than plain LSq (row 10) or \method{RANSAC} (row 11).

\begin{table}
  \centering
    \small
    \begingroup
    \setlength{\tabcolsep}{4pt} 
\begin{tabular}{cccccc || rr|rr}
  & & & & & & \multicolumn{2}{c}{P@5} & \multicolumn{2}{c}{P@15} \\
  & M & PW         & H     & W            & F            &   orig &   rean &   orig &   rean \\
\hline
(1) & R & -- & LSq & -- & \(\checkmark\) & 5.7 & 0.8 & 16.6 & 10.7\\
(2) & R & -- & LSq & \(\checkmark\) & \(\checkmark\) & 7.0 & 2.1 & 22.5 & 17.3\\
\hline
(3) & R & \(\checkmark\) & LSq & -- & \(\checkmark\) & 57.6 & 63.6 & 68.1 & 68.9\\
(4) & R & \(\checkmark\) & LSq & \(\checkmark\) & \(\checkmark\) & 66.7 & 74.3 & 75.5 & 76.4\\
\hline
(5) & R & C & LSq & -- & \(\checkmark\) & 73.1 & 82.1 & 89.9 & 92.0\\
\textbf{(6)} & \textbf{R} & \textbf{C} & \textbf{LSq} & \textbf{\(\checkmark\)} & \textbf{\(\checkmark\)} & \textbf{80.6} & \textbf{90.4} & \textbf{93.9} & \textbf{95.6}\\
\hline
(7) & R & C & LSq & \(\checkmark\) & -- & 75.1 & 83.0 & 87.3 & 87.8\\
(8) & R & C & IRLSq & \(\checkmark\) & \(\checkmark\) & 80.6 & 90.4 & 93.9 & 95.6\\
(9) & R & C & RSAC & -- & \(\checkmark\) & 79.5 & 88.8 & 92.7 & 93.5\\
\hline
(10) & L & C & LSq & -- & -- & 66.9 & 74.8 & 82.3 & 82.6\\
(11) & L & C & RSAC & -- & -- & 72.8 & 80.9 & 84.4 & 85.1\\
(12) & L & C & LSq & \(\checkmark\) & -- & 72.8 & 81.0 & 86.1 & 87.1\\
\end{tabular}
\endgroup

\caption{Ablation study on \dataset{POT-210}, evaluated on the original ground truth ({\it orig}) and the reannotation ({\it rean}).
  In all experiments, weighted least squares perform better than non-weighted alternative in both P@5 and P@15.
  M -- flow method: \method{RAFT} (R), \method{LiteFlowNet2} (L).
  PW -- use of the global pre-warped flow: never (--), always ($\checkmark$), controlled (C).
  H -- homography estimation method: least squares (LSq), iterative re-weighted least squares with Huber loss (IRLSq), \method{RANSAC} (RSAC).
  W -- using the estimated weights.
  F -- using the fine-tuned \method{RAFT} flow.
}
  \label{tab:ablation}
\end{table}

\subsection{Weights Evaluation}
\begin{figure}
  \centering
  \includegraphics[width=1\linewidth]{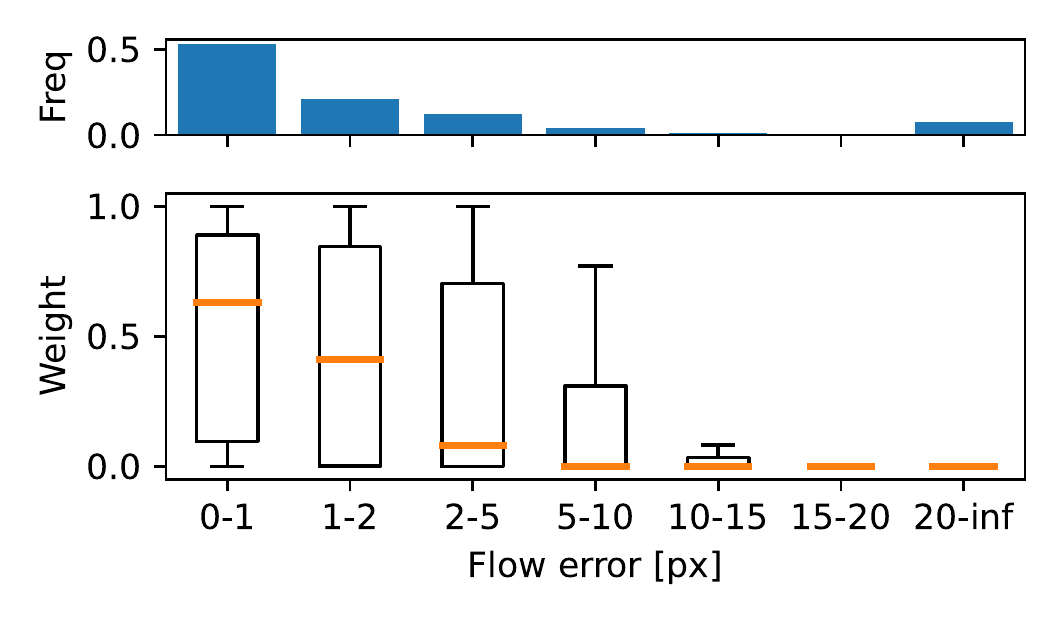}
  \caption{Weight distribution for different optical flow error ranges measured against the re-annotated \method{POT-210}~\cite{liang2017planar} ground truth.
    Median in orange.
    Top: frequency of each flow error range.
    The weight network learned to assign zero weight to incorrect flow vectors (outliers) and high weight to some correct flow vectors.
    }
  \label{fig:weight-distribution}
\end{figure}

Figure~\ref{fig:weight-distribution} shows how the learned weights correlate with the optical flow quality.
Low-textured areas and ambiguous features are often assigned a low weight (Fig.~\ref{fig:example-weights}) even when the corresponding optical flow is correct.
Importantly, the incorrect flow vectors are assigned low weights.

\subsection{POT-210 and POT-280 evaluation}
\begin{table}
  \centering
  \small
  \begingroup
  \setlength{\tabcolsep}{5pt} 
  \begin{tabular}{lrr | rr | rr}
    & & & \multicolumn{2}{c}{P@5} & \multicolumn{2}{c}{P@15} \\
method & year & FPS & orig & rean & orig & rean \\
\hline
GOP-ESM \cite{lin2019robust} & 2019 & 4.95\textsuperscript{*} & 42.9 & -- & 49.7 & --\\
SuperGlue \cite{sarlin2020superglue,liang2021planar} & 2020 & 3.7\textsuperscript{*} & 39.1 & 42.1 & 58.0 & 55.7\\
Gracker \cite{wang2017gracker} & 2017 & 4.8\textsuperscript{*} & 39.2 & -- & 63.2 & --\\
SiamESM \cite{chen2019learning} & 2019 & -- & 58.7 & -- & 66.2 & --\\
SOSNet \cite{tian2019sosnet,liang2021planar} & 2019 & 1.5\textsuperscript{*} & 56.6 & 60.9 & 69.9 & 67.0\\
SIFT \cite{lowe2004distinctive,liang2021planar} & 2004 & 0.8\textsuperscript{*} & 62.2 & 65.8 & 71.3 & 69.6\\
OBD \cite{matveichev2021mobile} & 2021 & 30\textsuperscript{*} & 48.4 & 54.3 & 79.3 & 79.2\\
LISRD \cite{pautrat2020online,liang2021planar} & 2020 & 7\textsuperscript{*} & 61.6 & 68.3 & 79.6 & 79.2\\
HDN \cite{zhan2021homography} & 2022 & 10.6\textsuperscript{*} & 61.3 & 70.9 & 91.5 & 92.4\\
\hline
WOFT\textsubscript{\(\downarrow\)3} (ours) &  & 19.2 & 68.9 & 80.5 & 91.2 & 92.3\\
\textbf{WOFT (ours)} &  & 3.5 & \textbf{80.6} & \textbf{90.4} & \textbf{93.9} & \textbf{95.6}\\
  \end{tabular}
  \endgroup
  \caption{Results on \dataset{POT-210}~\cite{liang2017planar} dataset.
  The proposed \method{WOFT} tracker sets a new state-of-the-art performance in both accuracy (P@5) and robustness (P@15).
  Evaluated on the original ground truth ({\it orig}) and the re-annotation ({\it rean}).
  Tracking speed in frames per second (FPS). \textsuperscript{*} speeds from the papers, different hardware.
}
  \label{tab:main-results}
\end{table}

\begin{figure*}
  \centering
  \includegraphics[page=2,width=0.32\linewidth]{POT210_merged_AE_roc} %
  \includegraphics[page=3,width=0.32\linewidth]{POT210_merged_AE_roc} %
  \includegraphics[page=4,width=0.32\linewidth]{POT210_merged_AE_roc} \\
  \includegraphics[page=5,width=0.32\linewidth]{POT210_merged_AE_roc} %
  \includegraphics[page=6,width=0.32\linewidth]{POT210_merged_AE_roc} %
  \includegraphics[page=7,width=0.32\linewidth]{POT210_merged_AE_roc} \\
  \includegraphics[page=8,width=0.32\linewidth]{POT210_merged_AE_roc} %
  \framebox{\includegraphics[page=1,width=0.32\linewidth]{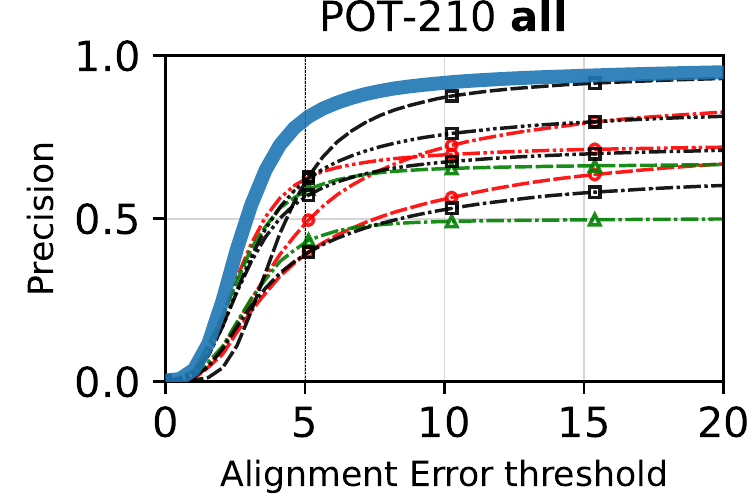}} %
  \hspace{1.4cm} %
  \includegraphics[width=0.165\linewidth]{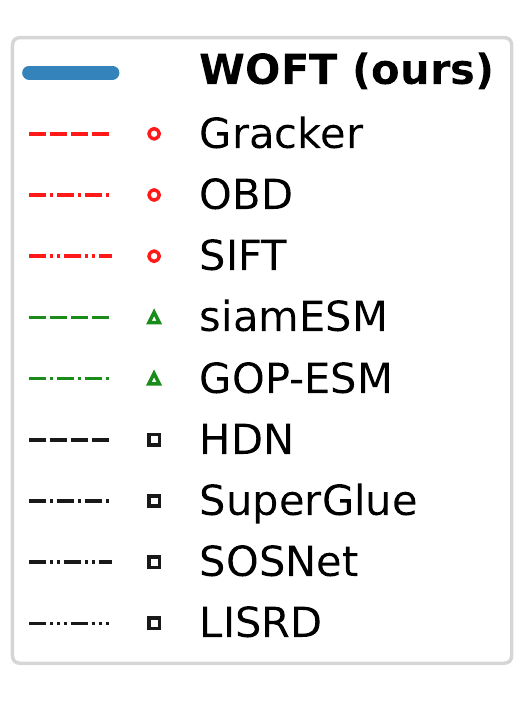} %
  \hspace{1.2cm} %
  \caption{Alignment Error on \dataset{POT-210}~\cite{liang2017planar} (original GT).
    \method{WOFT} performs well on all sequence types, reducing the error on the official 5px threshold to half of the best competitor.
    Method types: ({\it red circle}) -- keypoint, ({\it green triangle}) -- direct, ({\it black square}) -- deep. %
  }
  \label{fig:POT-results}
\end{figure*}
\begin{figure}
  \centering
    \includegraphics[page=1,width=0.6\linewidth]{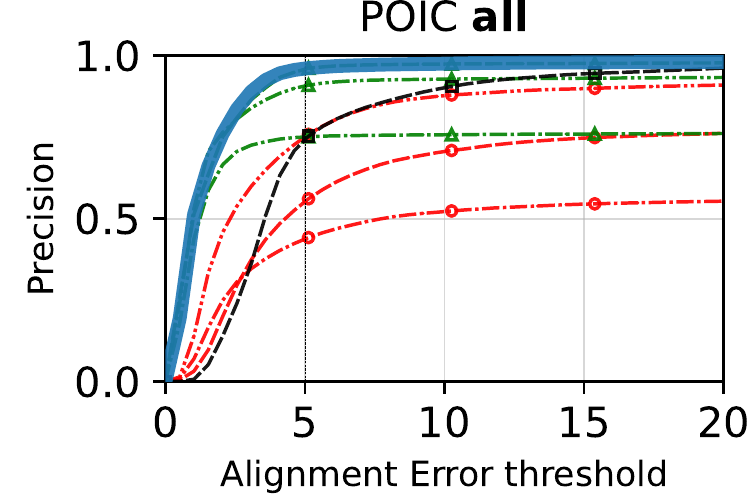} %
    \includegraphics[width=0.3\linewidth]{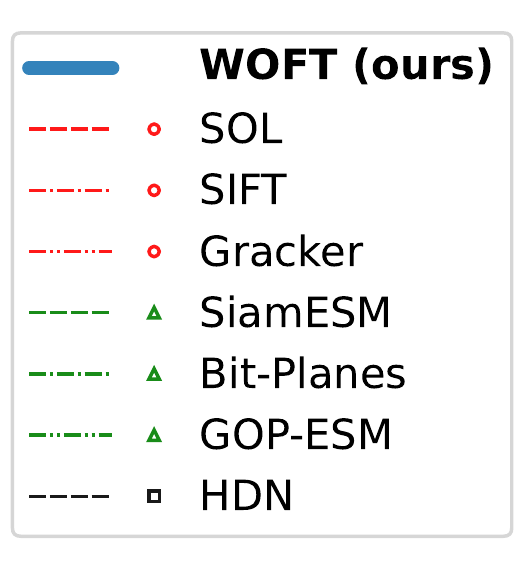}
    \caption{Alignment Error evaluation on \dataset{POIC}~\cite{lin2019robust}.
      The proposed \method{WOFT} achieves state-of-the-art with $96.1$ P@5 and $98.0$ P@15.}
    \label{fig:poic-plots}
\end{figure}

We compare \method{WOFT} method against the best performing methods on the \dataset{POT-210}~\cite{liang2017planar} dataset.  Namely keypoint methods:
\method{SIFT}~\cite{lowe2004distinctive},
\method{OBD}~\cite{matveichev2021mobile},
and \method{Gracker}~\cite{wang2017gracker},
deep control point regression \method{HDN}~\cite{zhan2021homography},
the deep learning based methods evaluated in \cite{liang2021planar}:
\method{SOSNet}~\cite{tian2019sosnet},
\method{SuperGlue}~\cite{sarlin2020superglue},
\method{LISRD}~\cite{pautrat2020online},
the direct methods:
\method{GOP-ESM}~\cite{lin2019robust},
and \method{Siam-ESM}~\cite{chen2019learning} (deep + direct).

The proposed \method{WOFT} achieves state-of-the-art on the \dataset{POT-210} dataset.
The Alignment Error $e_{AL}$ results are depicted on Fig.~\ref{fig:POT-results} and in Tab.~\ref{tab:main-results}.
Evaluated over all 210 sequences (\emph{all} plot) The \method{WOFT} tracker performs better than all the other methods, both in terms of accuracy (P@5), and robustness (P@15).
More than half of the 5px threshold errors of \method{WOFT} are explained by imprecise GT.
The \method{WOFT}\textsubscript{$\downarrow$3} variant operating on \(H/3 \times W/3\) images runs close to real-time and achieves state-of-the-art accuracy (details in supplementary).
\method{WOFT} also achieves top results on \dataset{POT-280}~\cite{liang2021planar} ($76.9$ P@5, $93.2$ P@15), see supplementary.

\begin{figure*}
  \centering
  \includegraphics[trim=0 100 0 0,clip,width=0.19\textwidth]{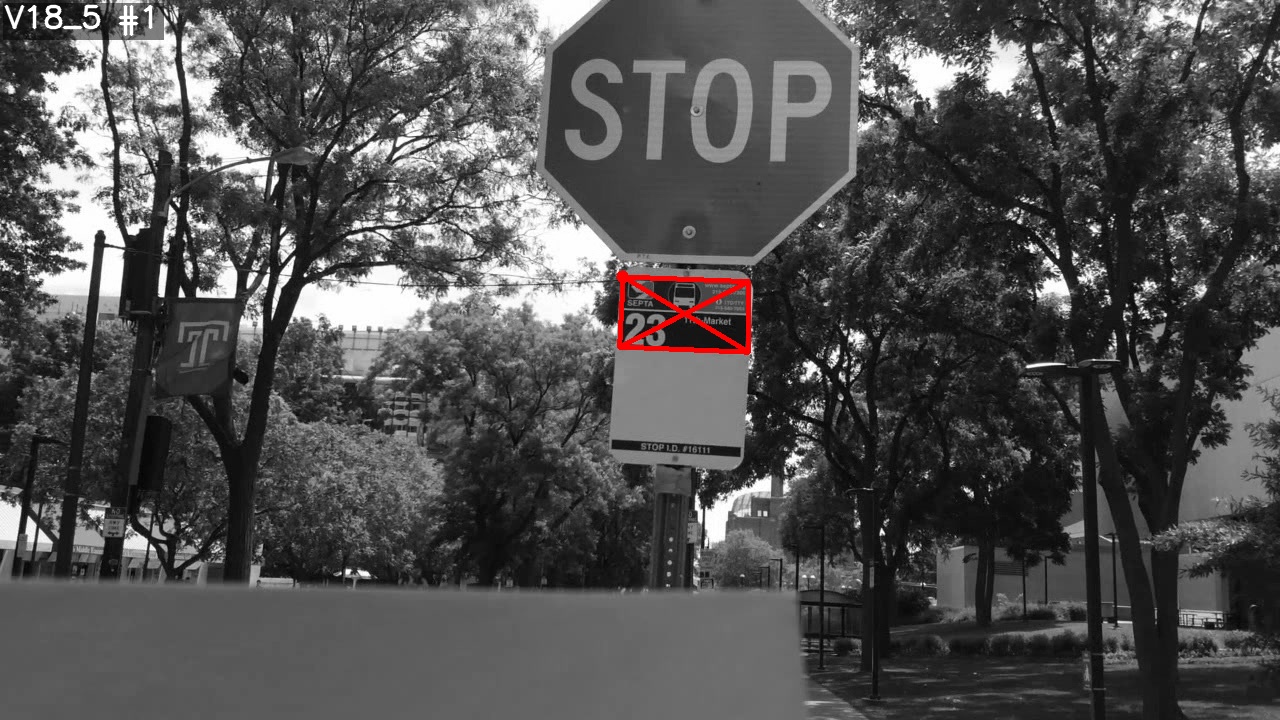} %
  \includegraphics[trim=0 100 0 0,clip,width=0.19\textwidth]{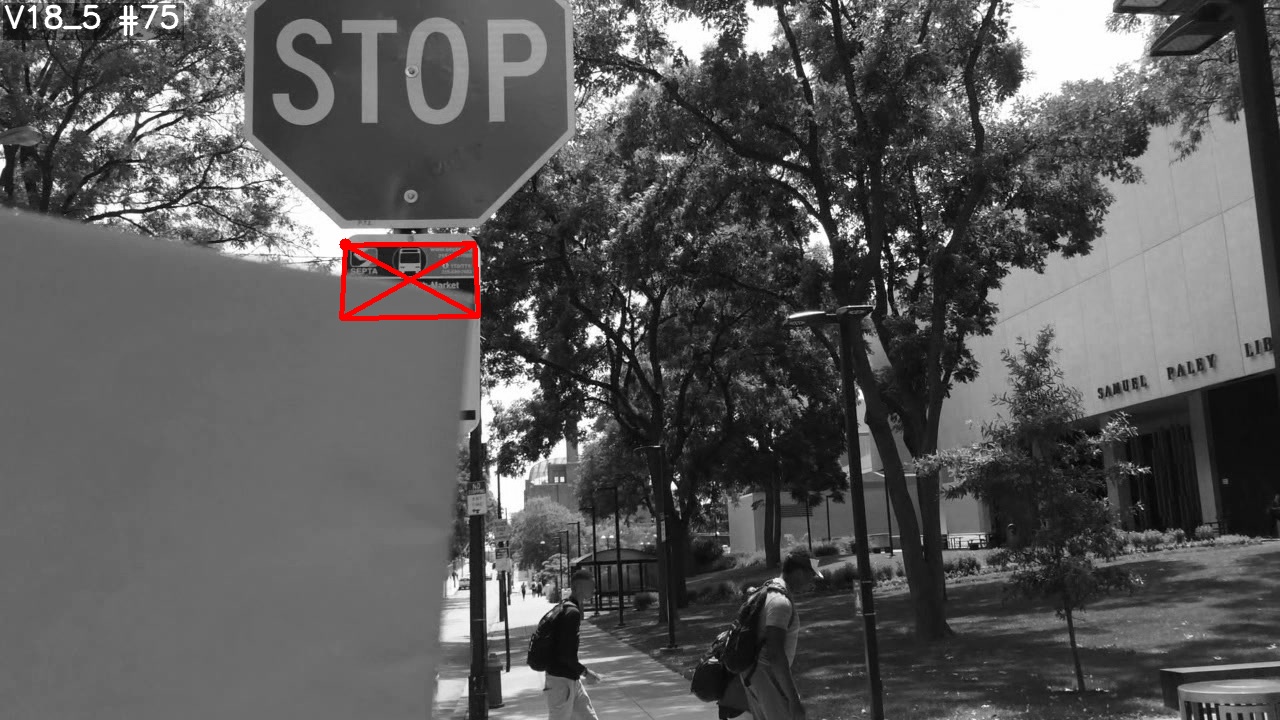} %
  \includegraphics[trim=0 100 0 0,clip,width=0.19\textwidth]{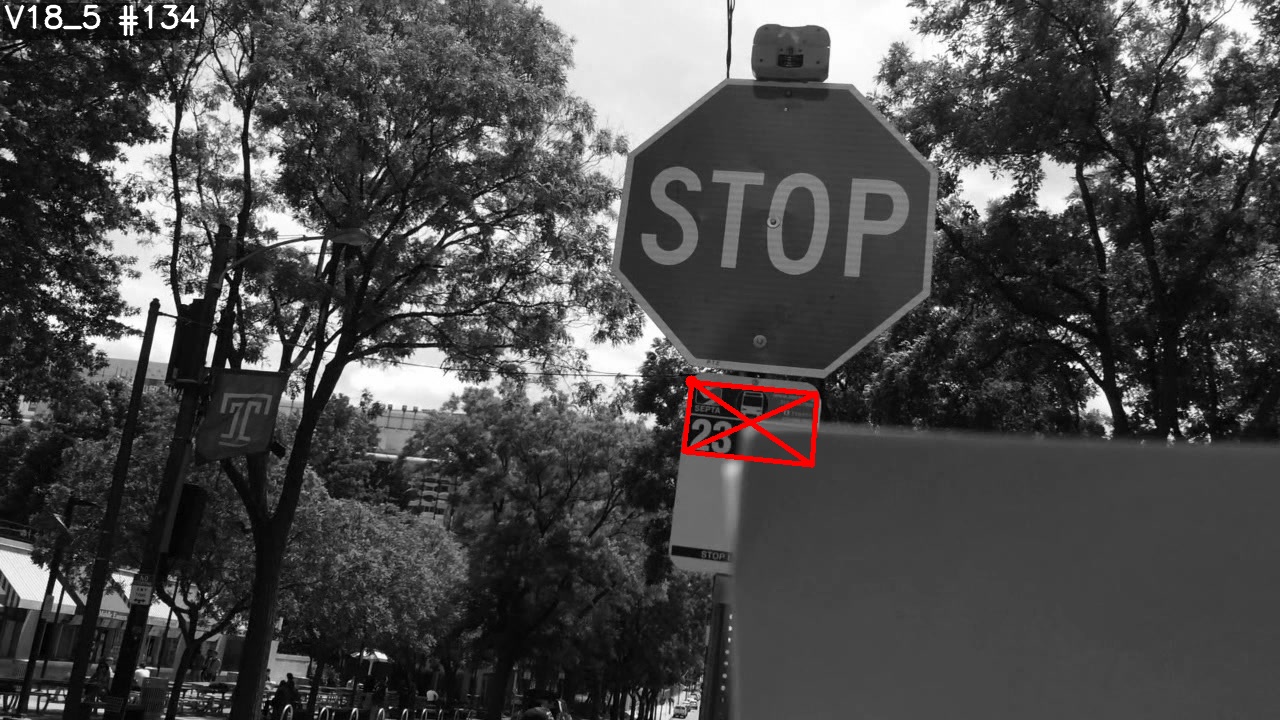} %
  \includegraphics[trim=0 100 0 0,clip,width=0.19\textwidth]{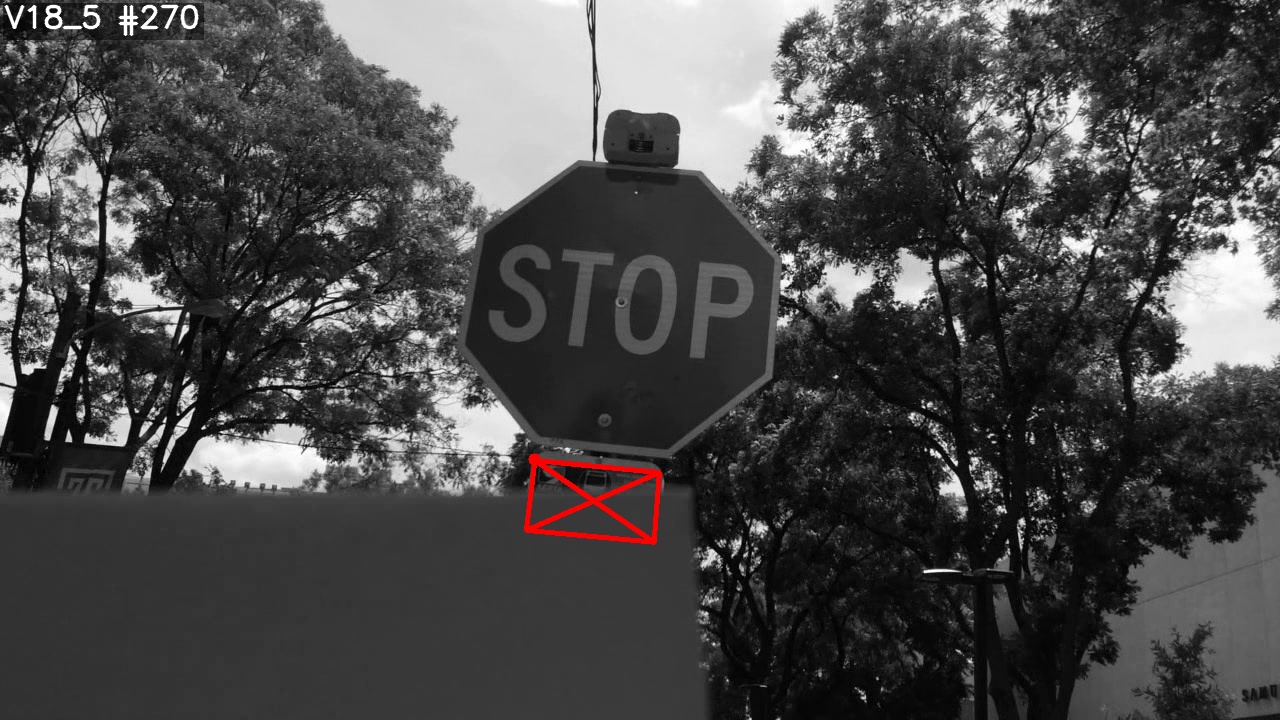} %
  \includegraphics[trim=0 100 0 0,clip,width=0.19\textwidth]{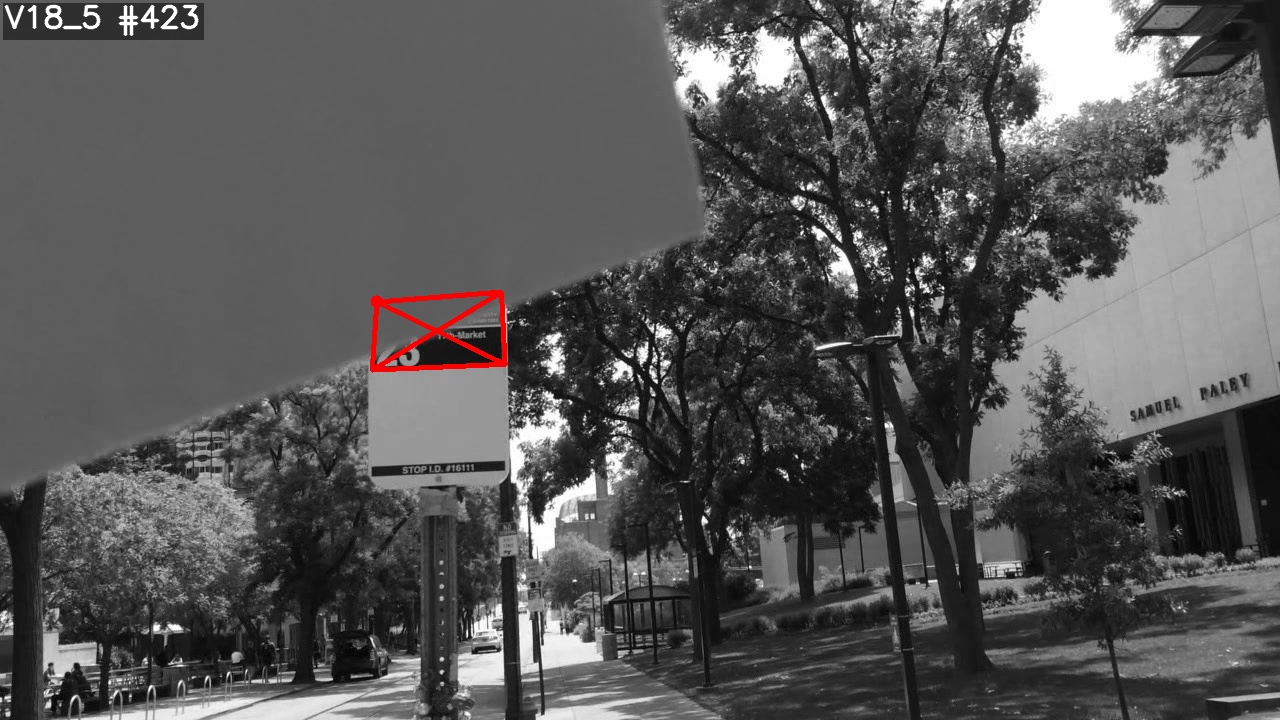}

  \includegraphics[width=0.19\textwidth]{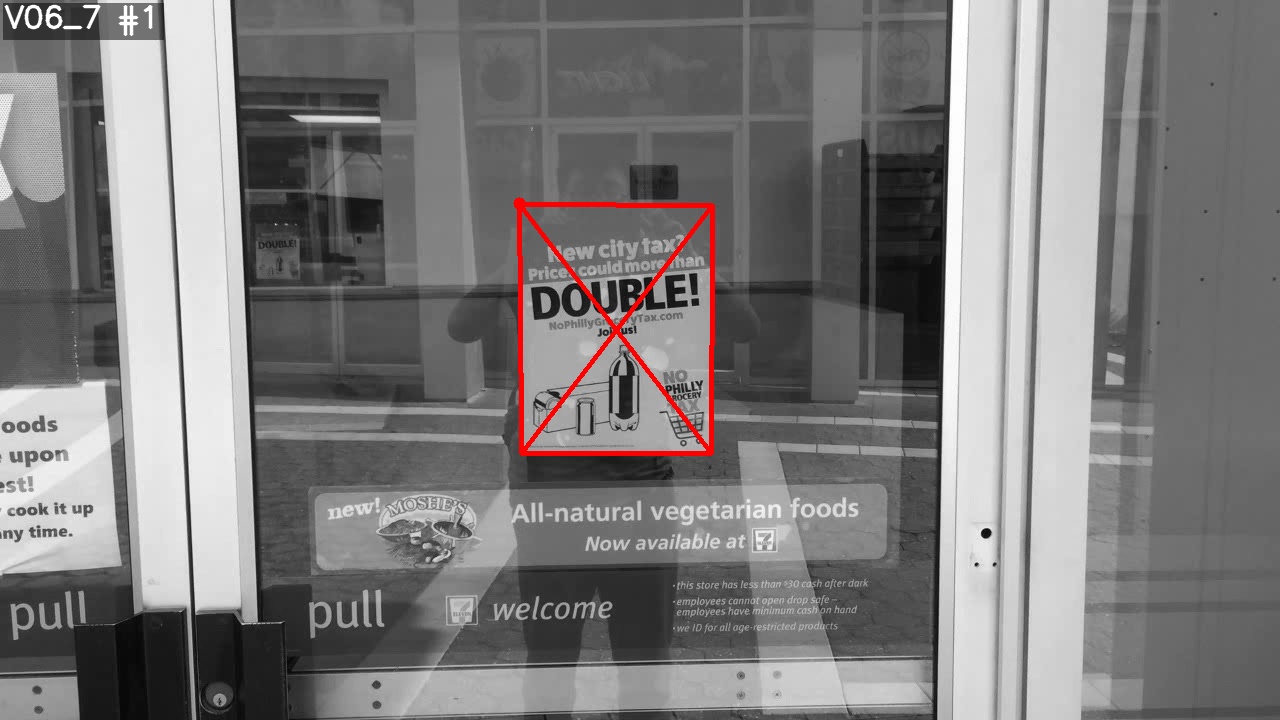} %
  \includegraphics[width=0.19\textwidth]{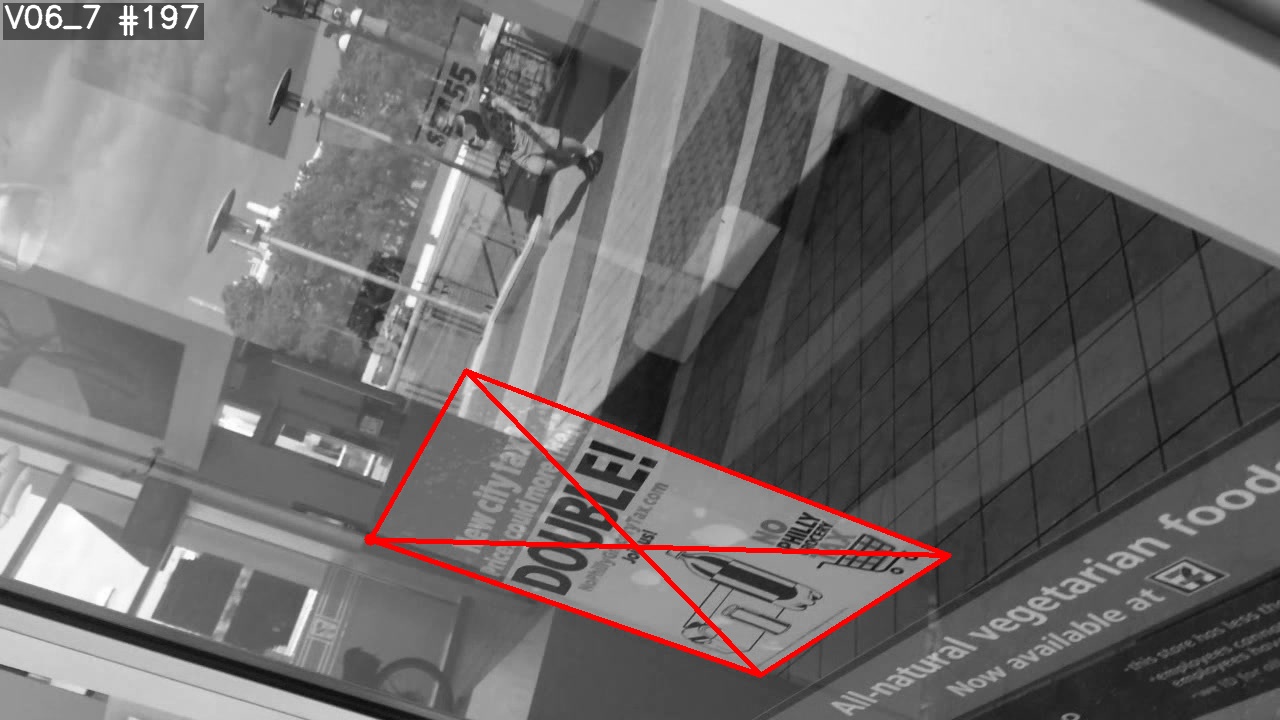} %
  \includegraphics[width=0.19\textwidth]{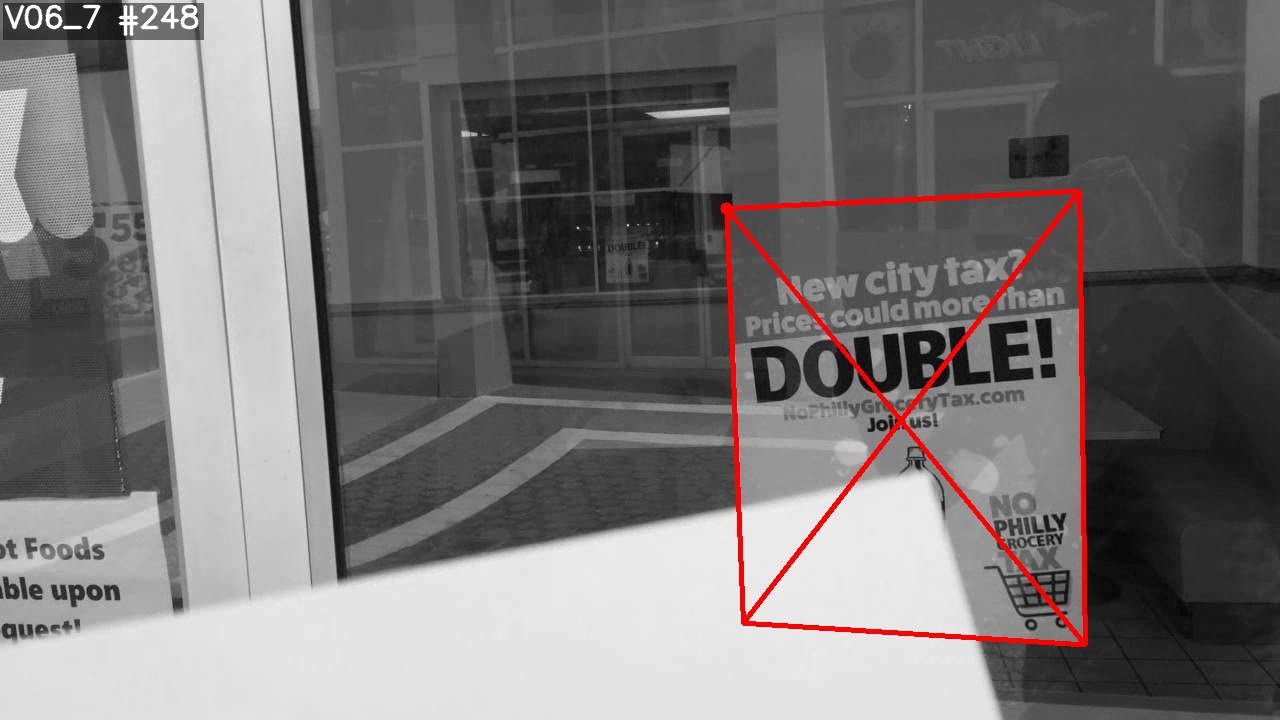} %
  \includegraphics[width=0.19\textwidth]{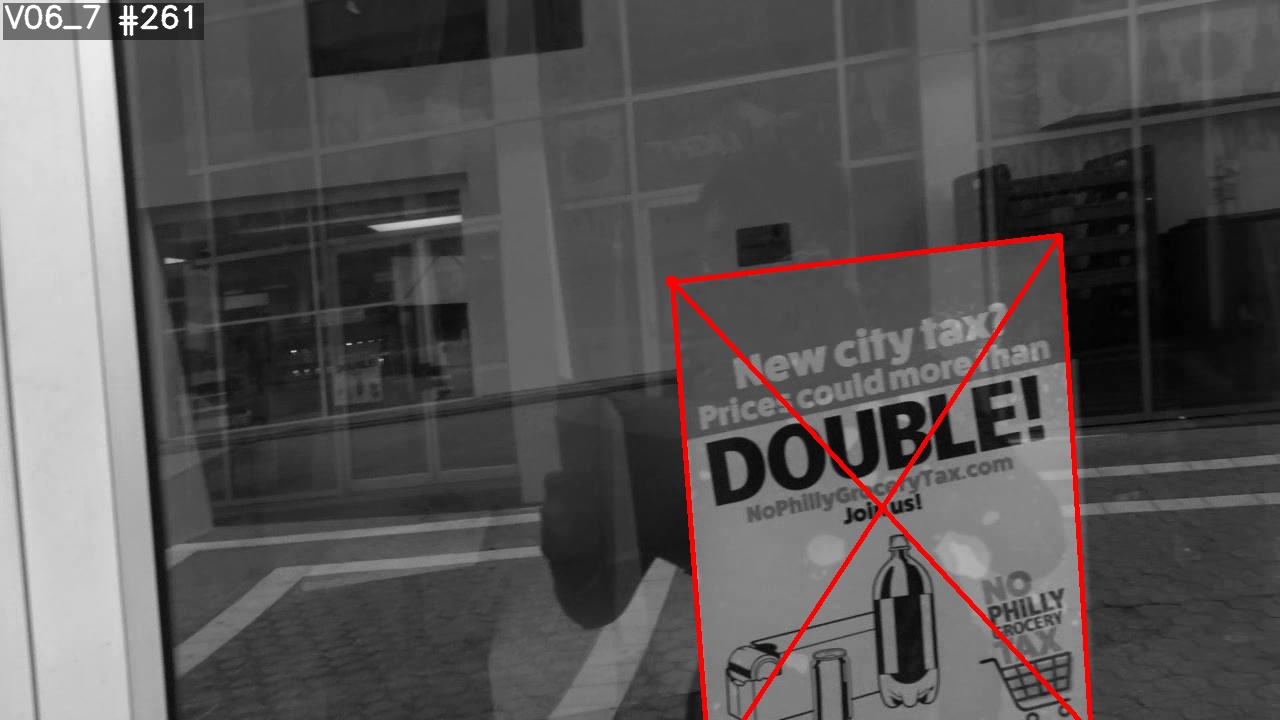} %
  \includegraphics[width=0.19\textwidth]{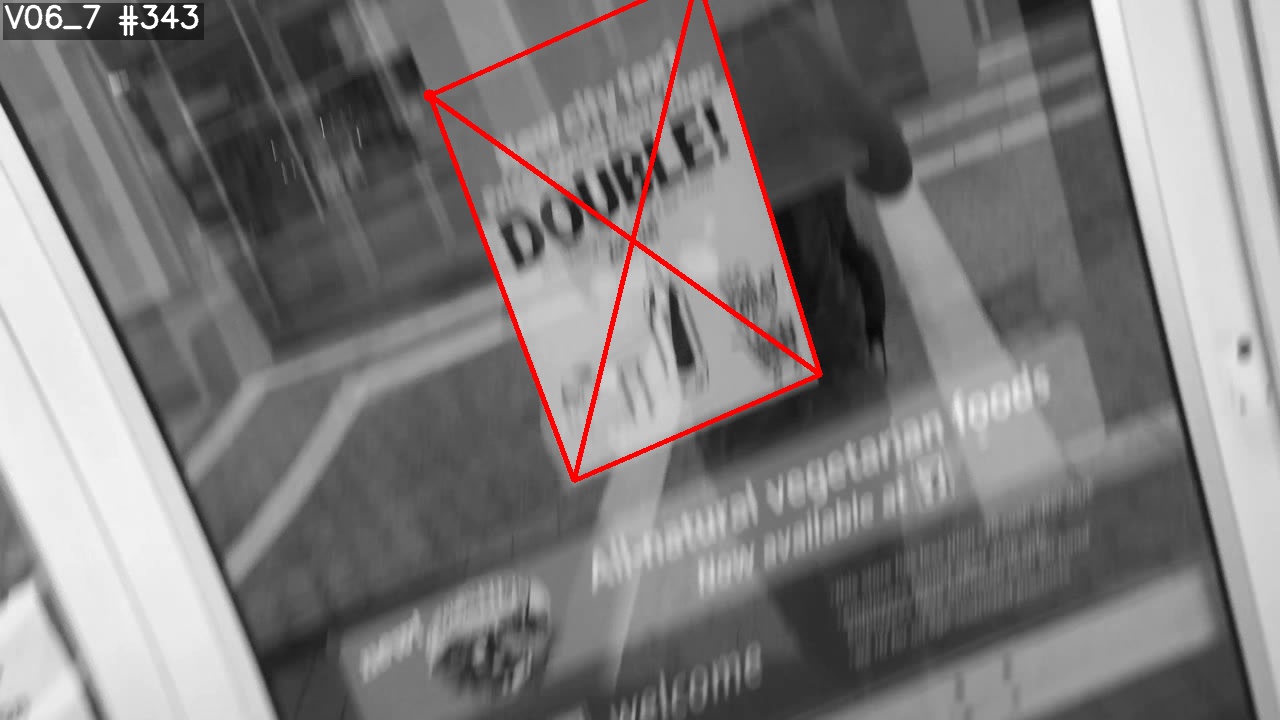}

  \includegraphics[width=0.19\textwidth]{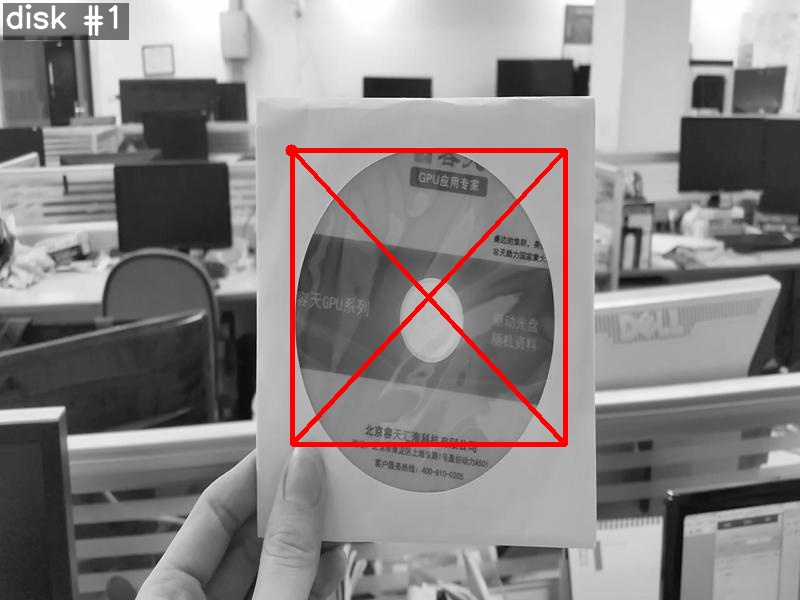} %
  \includegraphics[width=0.19\textwidth]{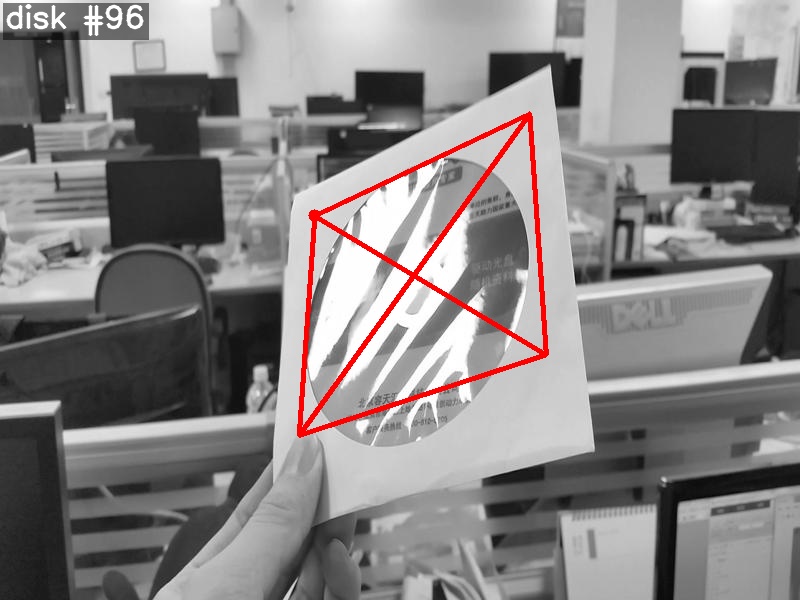} %
  \includegraphics[width=0.19\textwidth]{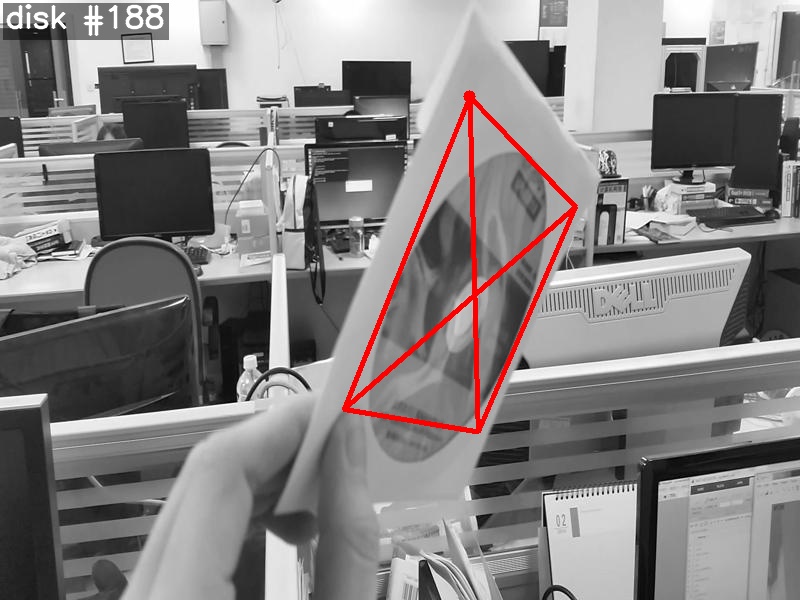} %
  \includegraphics[width=0.19\textwidth]{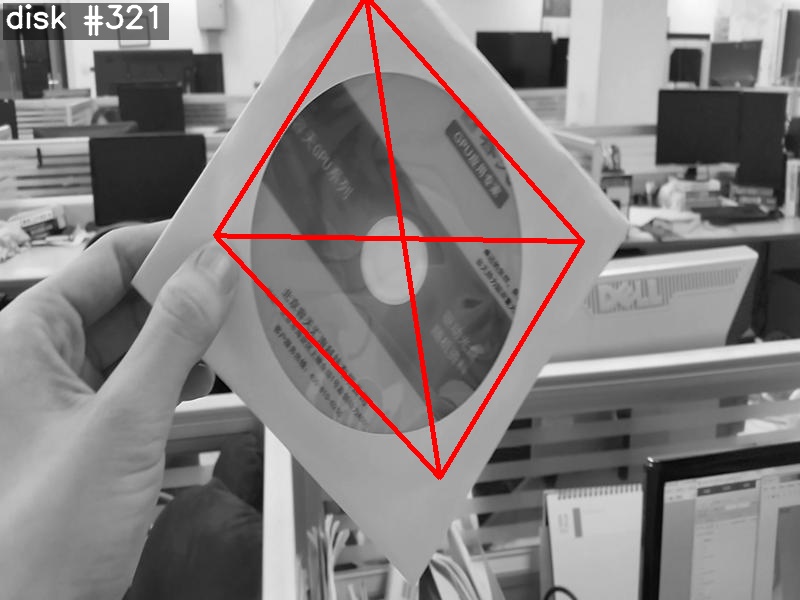} %
  \includegraphics[width=0.19\textwidth]{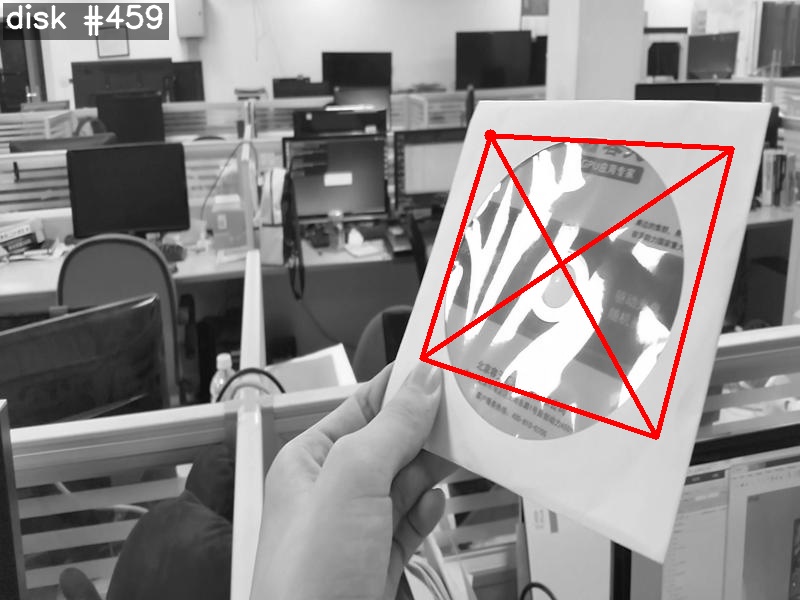}

  \includegraphics[trim=0 50 0 0,clip,width=0.19\textwidth]{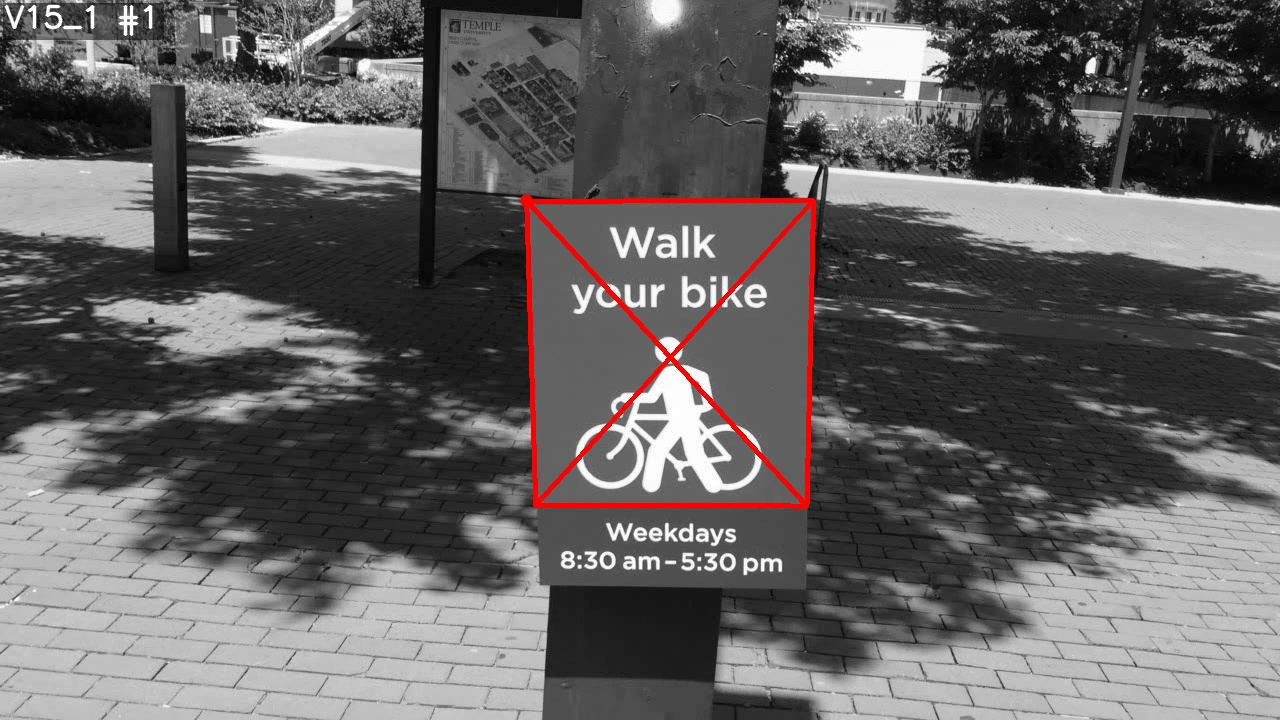} %
  \includegraphics[trim=0 50 0 0,clip,width=0.19\textwidth]{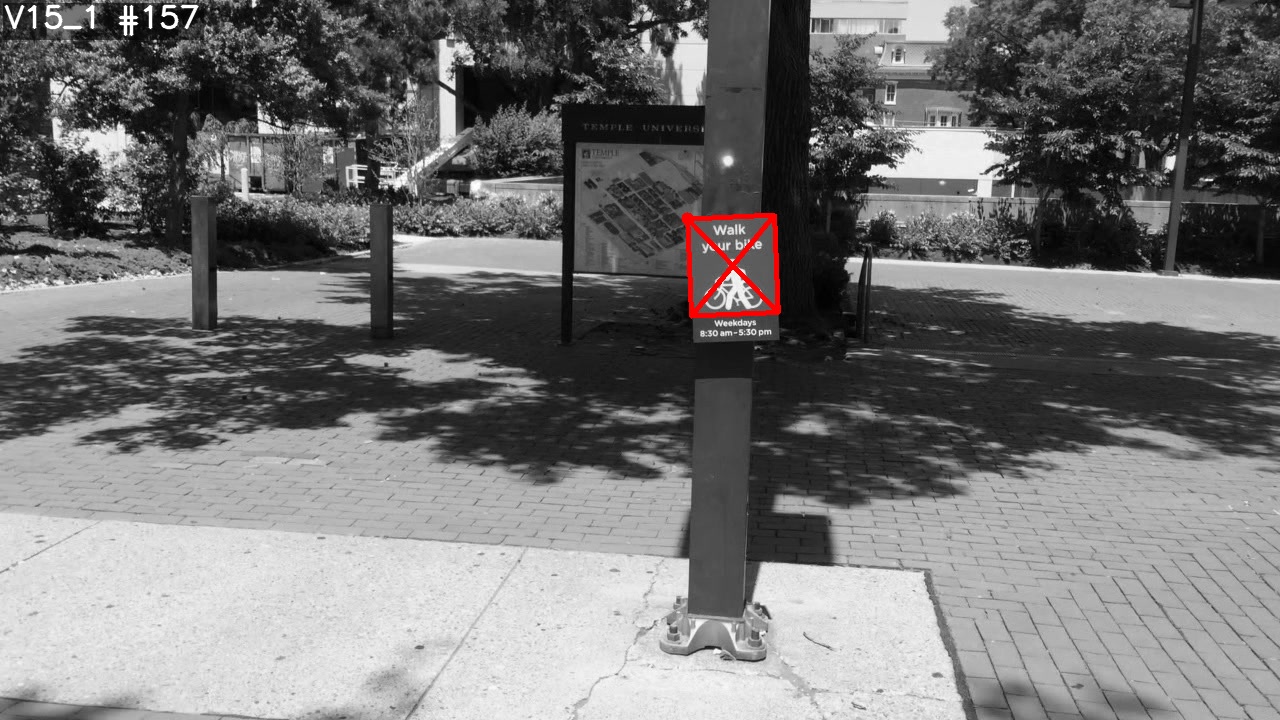} %
  \includegraphics[trim=0 50 0 0,clip,width=0.19\textwidth]{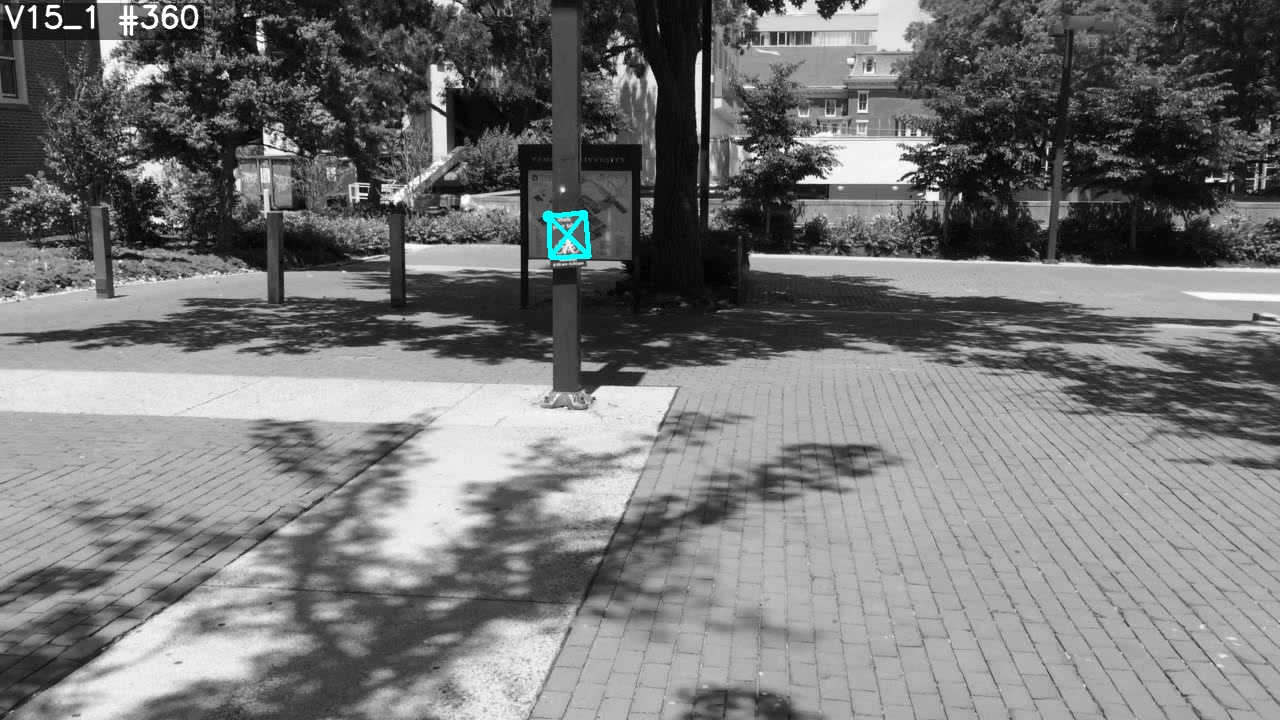} %
  \includegraphics[trim=0 50 0 0,clip,width=0.19\textwidth]{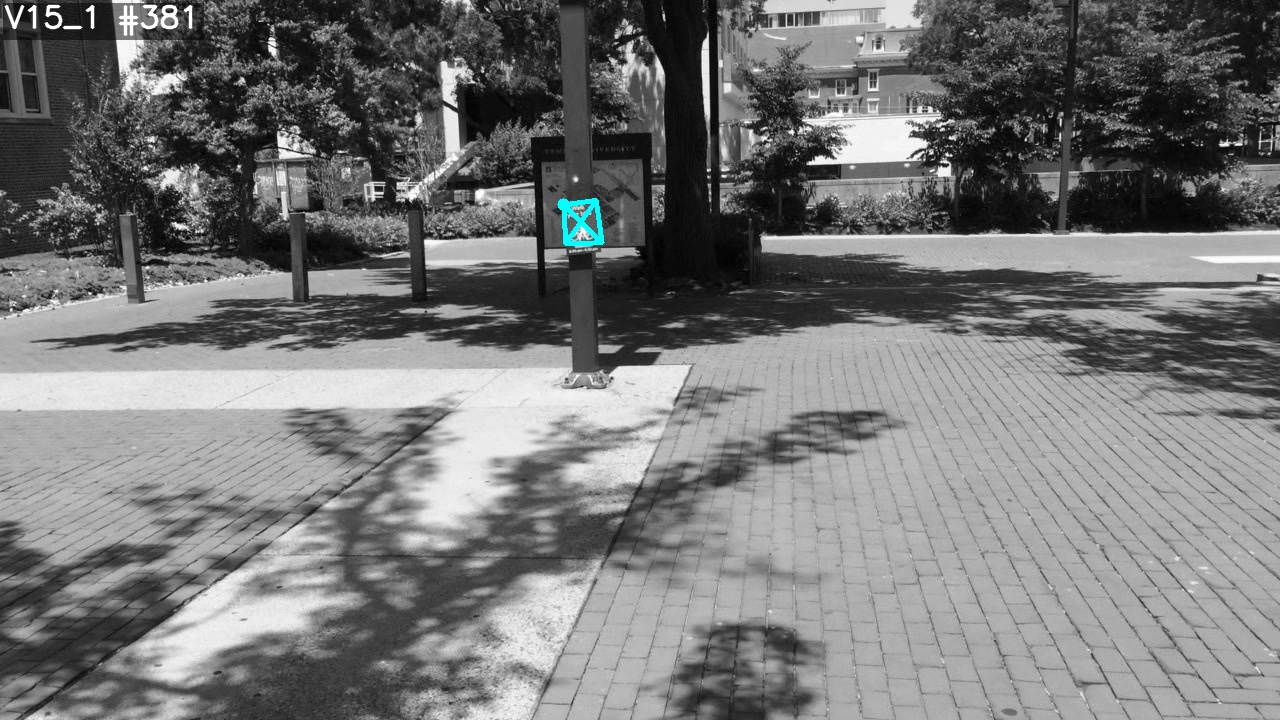} %
  \includegraphics[trim=0 50 0 0,clip,width=0.19\textwidth]{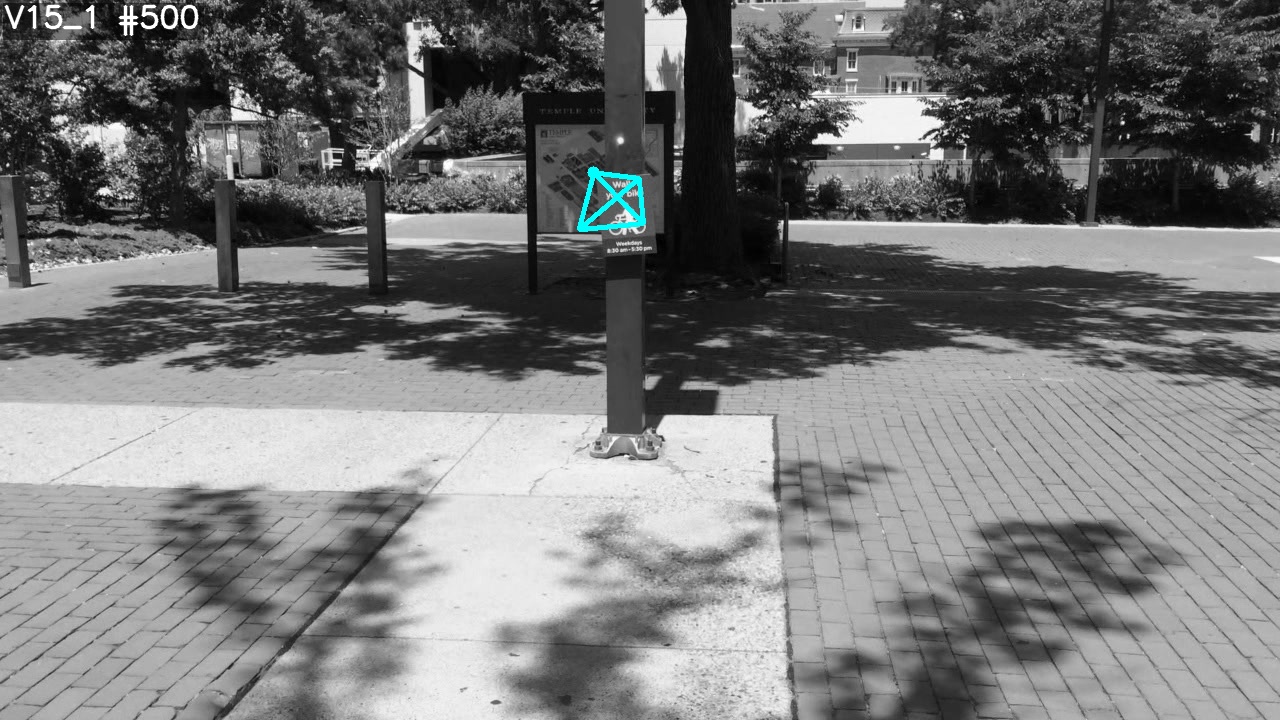}

  \includegraphics[width=0.19\textwidth]{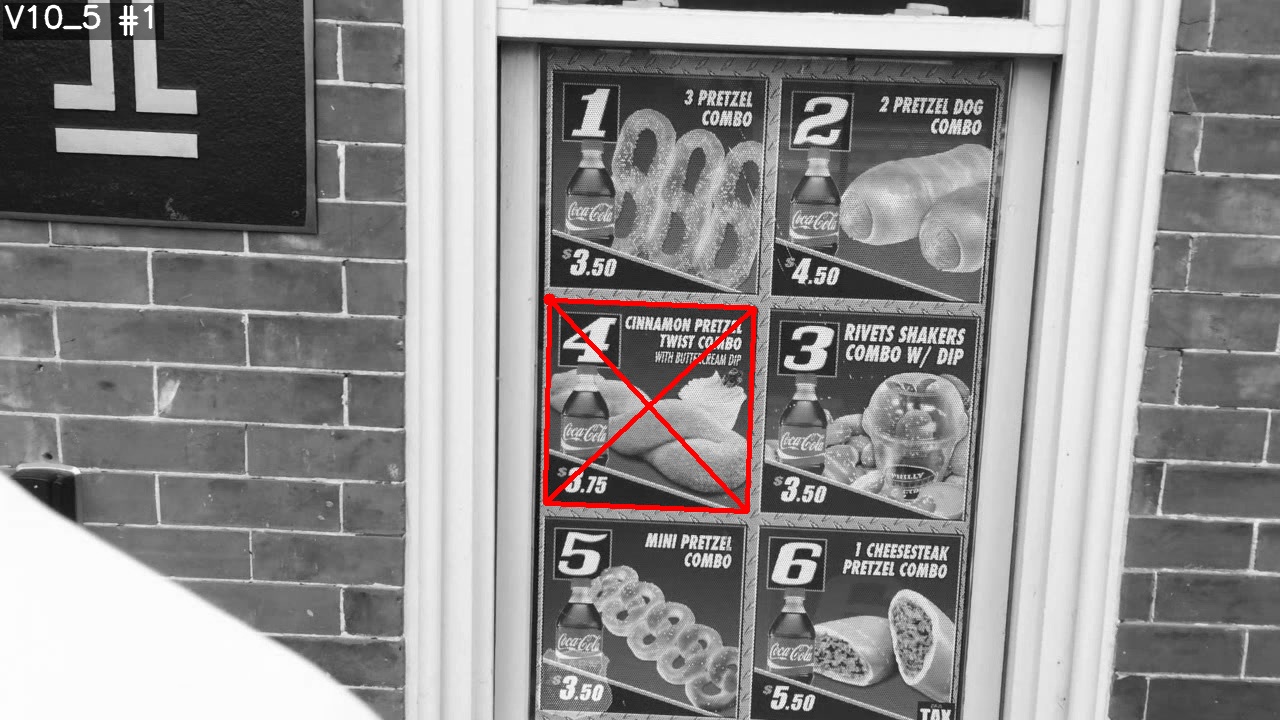} %
  \includegraphics[width=0.19\textwidth]{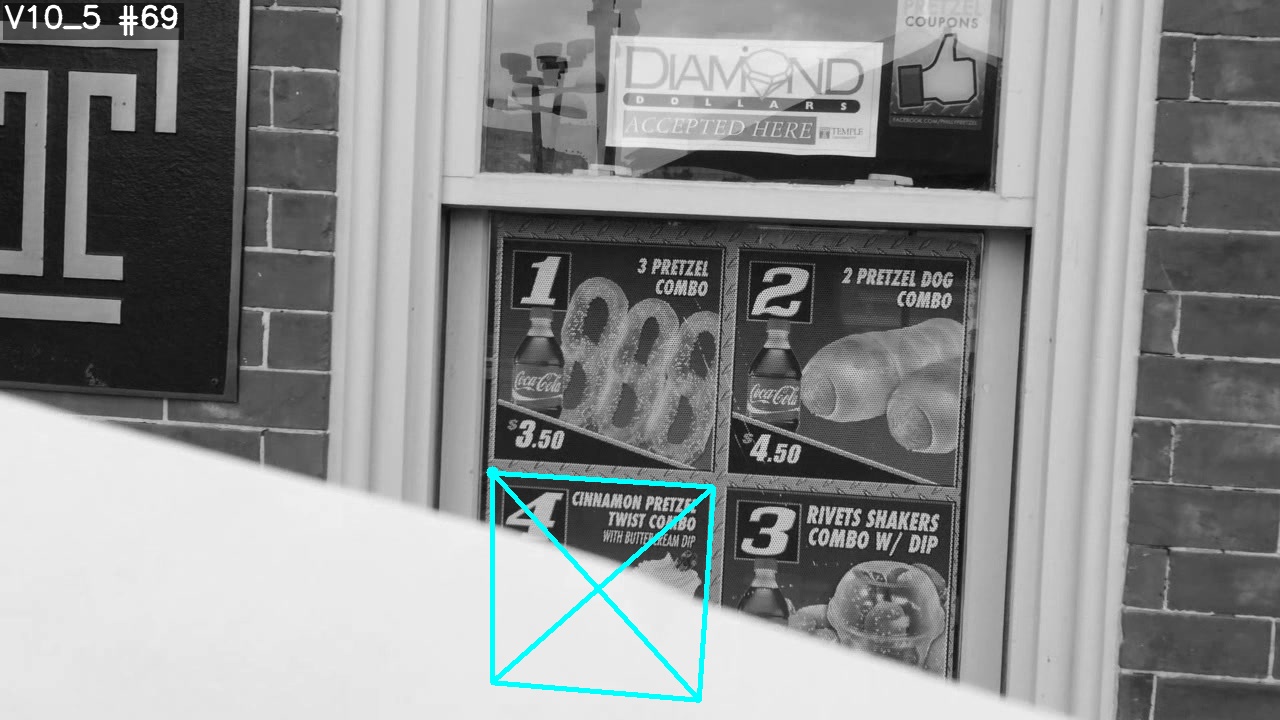} %
  \includegraphics[width=0.19\textwidth]{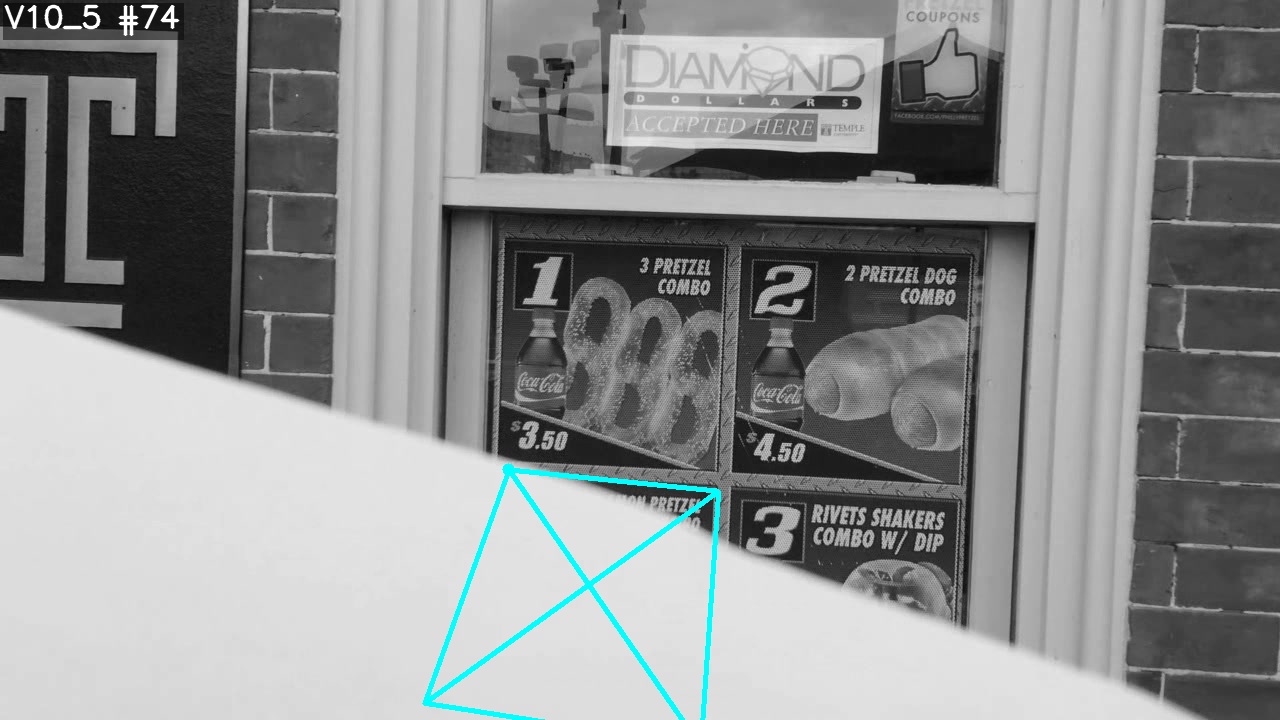} %
  \includegraphics[width=0.19\textwidth]{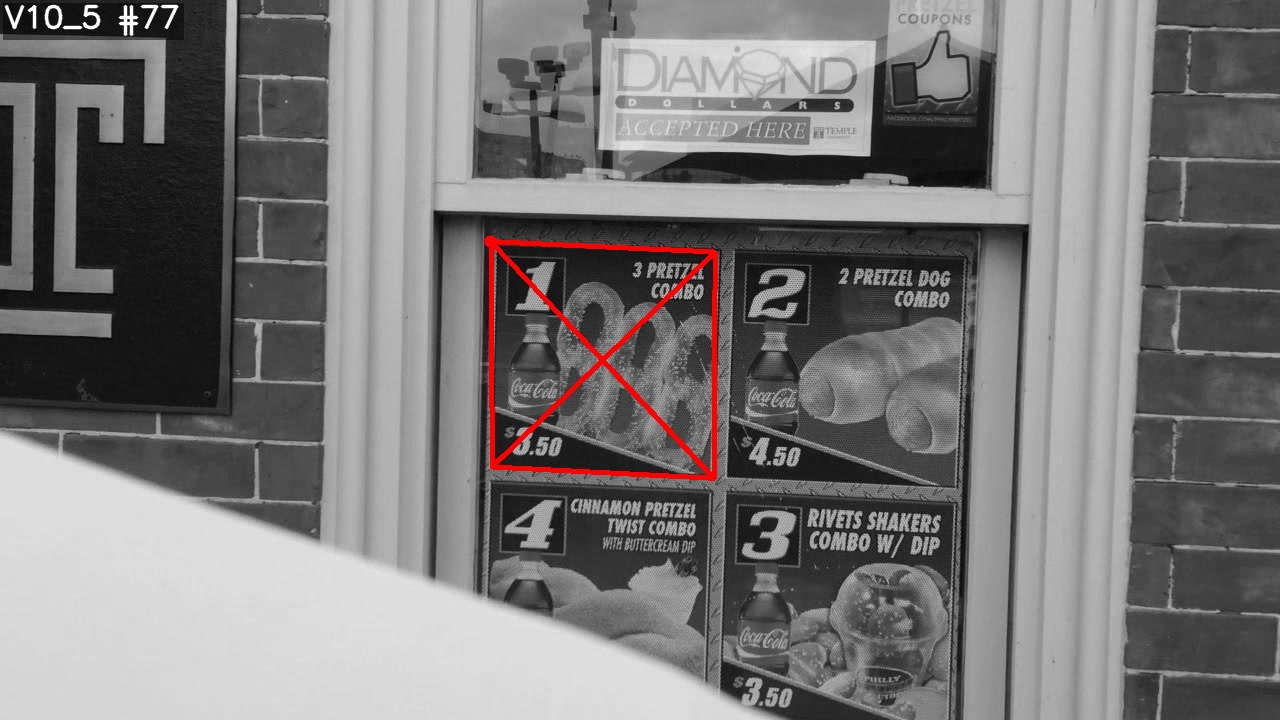} %
  \includegraphics[width=0.19\textwidth]{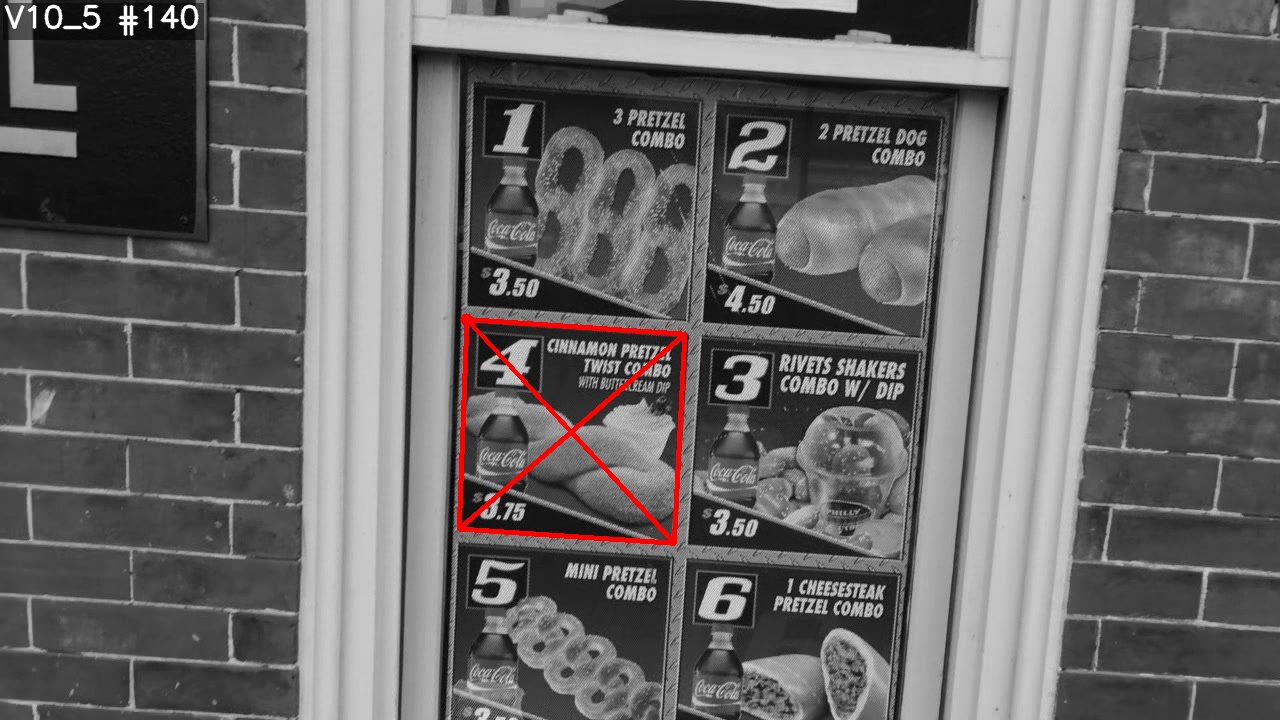}

  \includegraphics[trim=0 270 0 0,clip,width=0.19\textwidth]{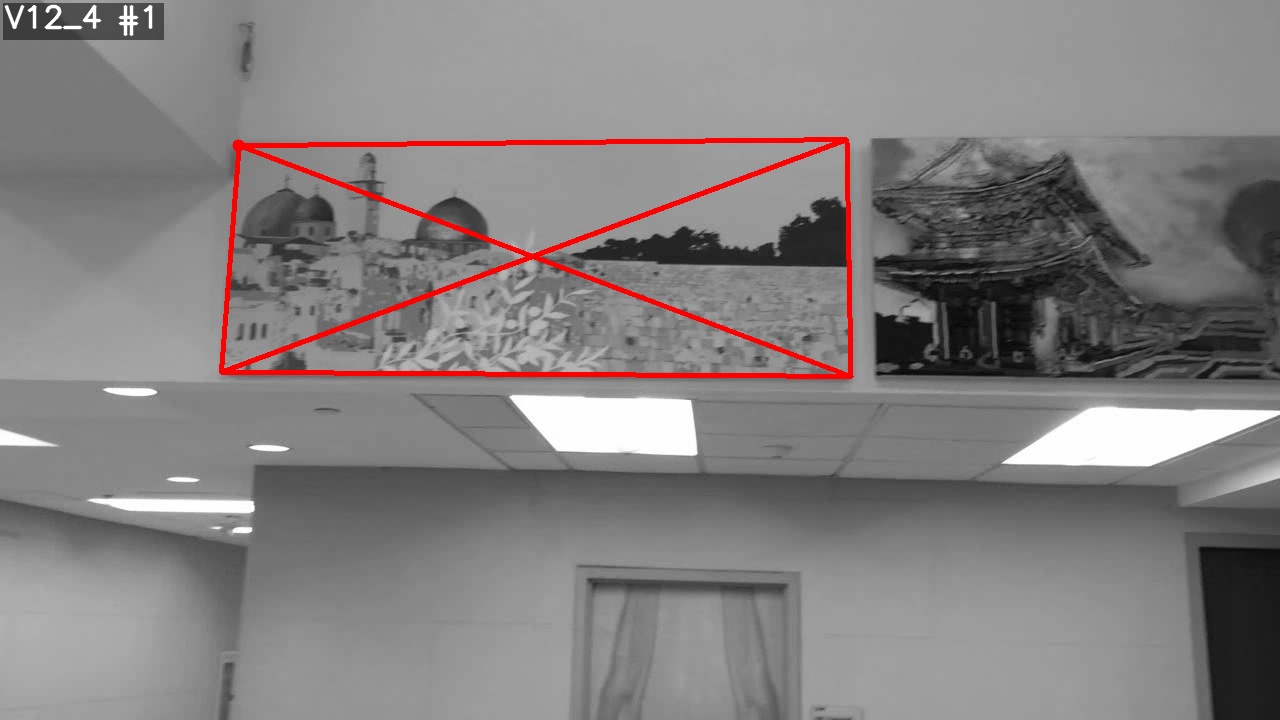} %
  \includegraphics[trim=0 270 0 0,clip,width=0.19\textwidth]{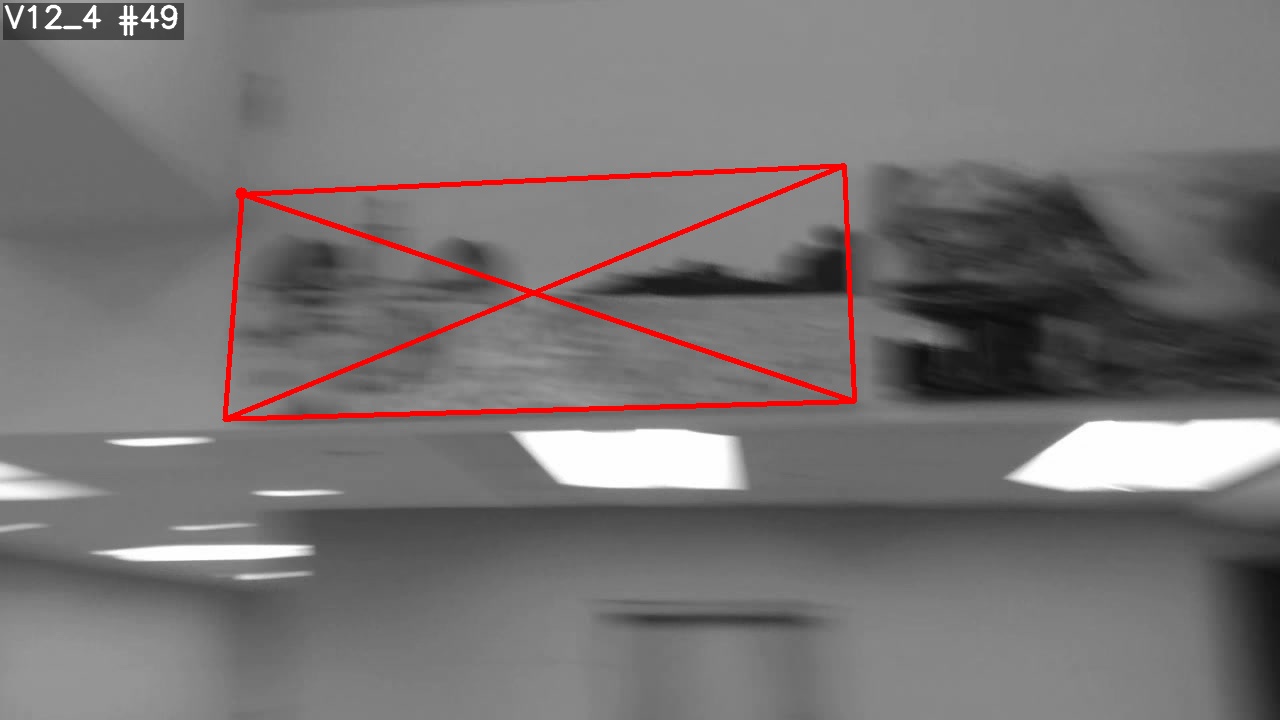} %
  \includegraphics[trim=0 270 0 0,clip,width=0.19\textwidth]{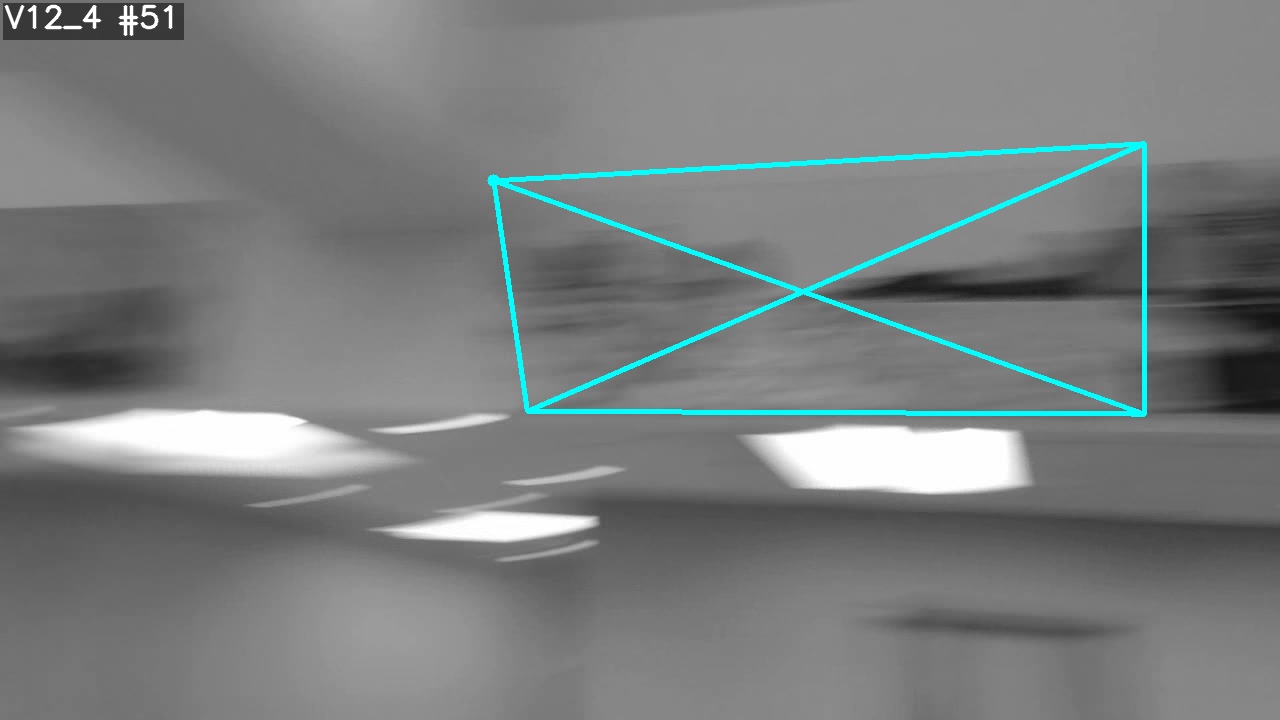} %
  \includegraphics[trim=0 270 0 0,clip,width=0.19\textwidth]{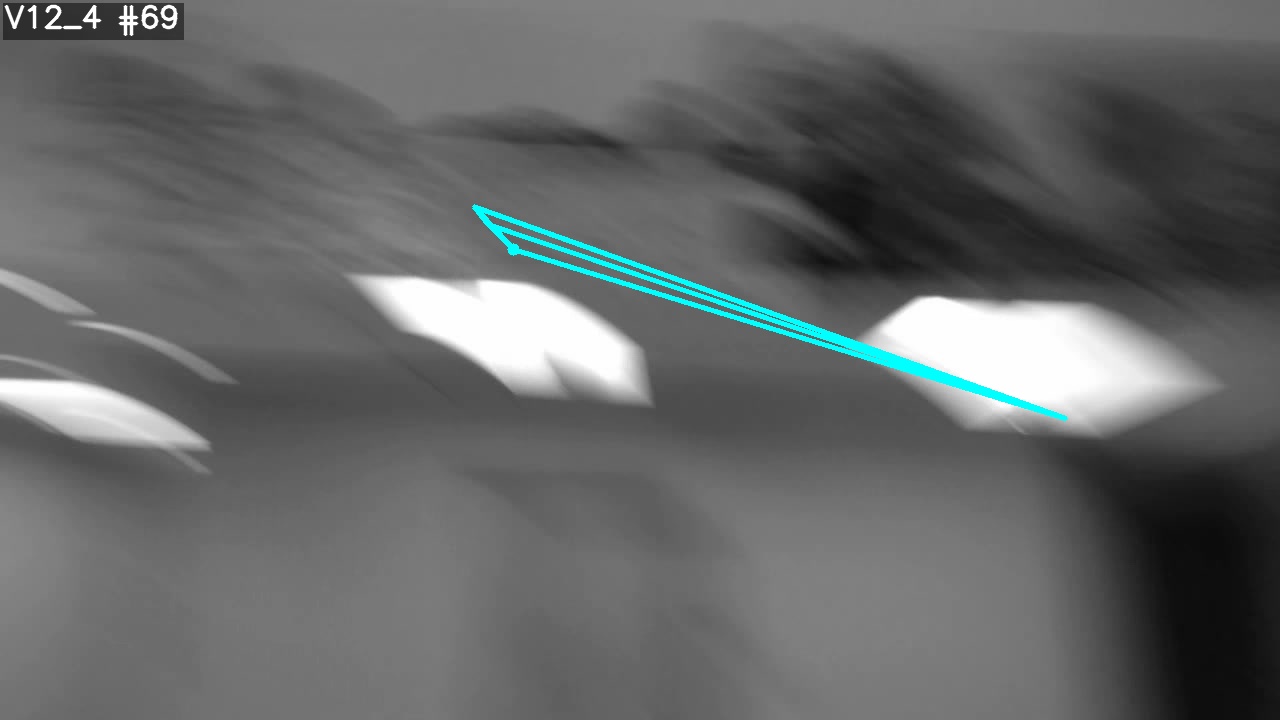} %
  \includegraphics[trim=0 270 0 0,clip,width=0.19\textwidth]{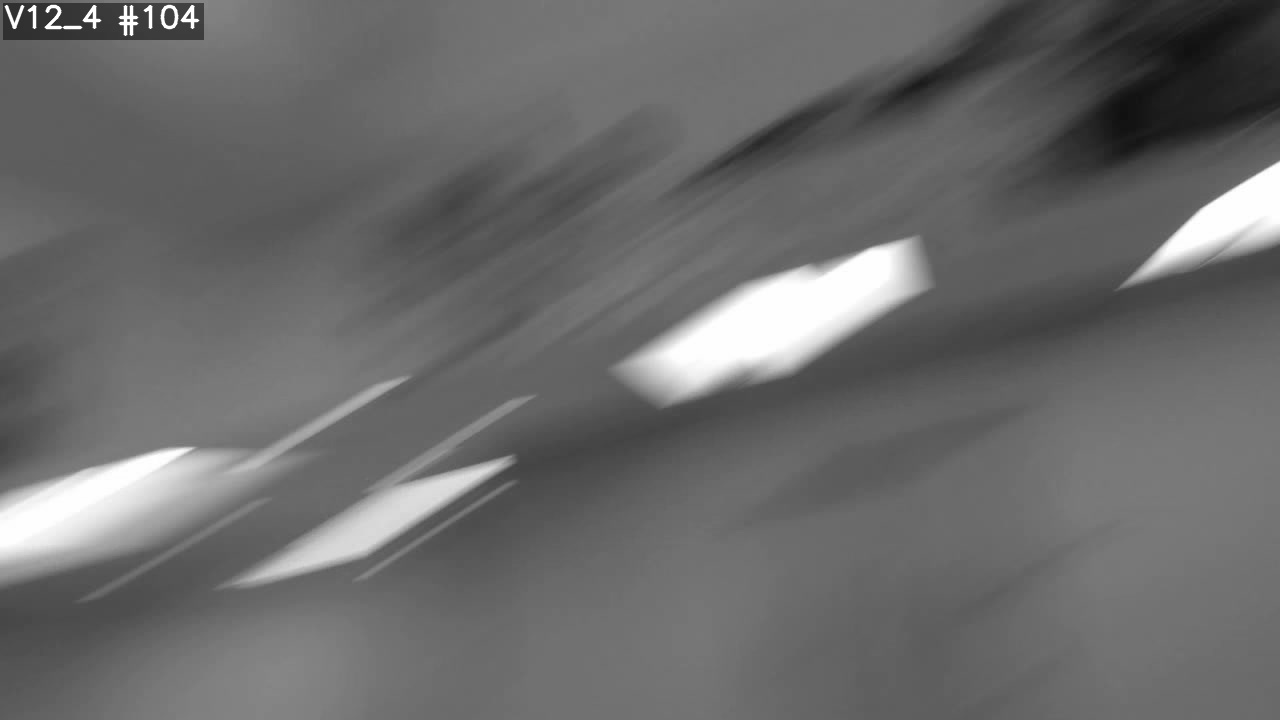}
  \caption{\method{WOFT} tracking. State visualization -  red: tracking, cyan: lost - switch to local flow. First row: \method{WOFT} handles strong occlusions on the \dataset{POT-210} V18\_5 sequence.
    Second row: successful tracking on the V06\_7 \dataset{POT-210} \emph{unconstrained} with perspective change, partial occlusion, scale change and motion blur.
    Third row: successful tracking in a \dataset{POIC} disk sequence, where a large part of the target surface changes appearance because of specular reflection.
    The last two rows show selected tracking failures.
    Row 4: the tracker is `lost` and did not recover because of a big scale difference w.r.t. the template frame, however, the local homography estimation prevents complete failure. 
    Row 5: the target becomes almost fully occluded and the tracker switches to track a nearby distractor patch.
    Later \method{WOFT} reacquired the correct target.
    Last row: \method{WOFT} can handle a moderate amount of motion blur, but fails on extremely blurred frames.
  }
  \label{fig:examples}
\end{figure*}
\subsection{POIC evaluation}
We compare (Fig.~\ref{fig:poic-plots}) the \method{WOFT} tracker performance with the top methods evaluated on the \dataset{POIC}~\cite{lin2019robust} dataset.
Apart from the methods evaluated on \dataset{POT-210}, this includes \method{SOL}~\cite{hare2012efficient} and \method{Bit-Planes}~\cite{alismail2016robust}.
\method{WOFT} achieves state-of-the-art results with $96.1$ P@5 and $98.0$ P@15.
More results are in supplementary materials. 
See Fig.~\ref{fig:examples} for \method{WOFT} output examples on both \method{POT-210} and \method{POIC}.

\section{Discussion and Limitations}
The \method{WOFT} tracker handles partial occlusions, a moderate amount of motion blur, and the illumination changes and lack of texture present in the \dataset{POIC} dataset.
In comparison, other methods performing well on \dataset{POIC} (\method{SiamESM}~\cite{chen2019learning}, \method{GOP-ESM}~\cite{lin2019robust}) have low performance on \dataset{POT-210} and vice versa (\method{LISRD}~\cite{pautrat2020online}, \method{SIFT}~\cite{lowe2004distinctive}).
\method{WOFT} does not feature a re-detection scheme and estimates only the residual transformation after the pre-warp step.
This causes issues when the tracker gets lost for more than 10 frames on the \emph{scale} subset.
After resetting the pre-warp source frame to $G = 0$ (pre-warp with an identity homography), the scale component of the residual transformation is sometimes bigger than what the flow network can handle (see Fig.~\ref{fig:examples}).

We tested the proposed \method{WFH} homography method on the \method{RAFT} OF network, which is accurate (Fig.~\ref{fig:weight-distribution}), but slow (275ms per frame).
However, the OF estimation is an active area of research and we expect new accurate and fast methods to be published in the future.
The core idea of \method{WFH} -- flow weights computed from an OF cost-volume and a differentiable homography estimation with weighted LSq -- is applicable to other OF methods.
The ablation study results with \method{RAFT} replaced by \method{LiteFlowNet2} support this claim.
We also proposed a simple \method{WOFT}\textsubscript{$\downarrow\! 3$} variant that operates fast (19.2 FPS) and still achieves state-of-the-art.

\section{Conclusions}
We proposed a novel formulation of deep homography estimation by weighted least squares.
The weighted flow homography (\method{WFH}) module is differentiable and can be trained end-to-end together with an optical flow network that provides dense correspondences.
A novel planar object tracker, called \method{WOFT}, that uses \method{WFH} was evaluated on two complementary planar object tracking benchmarks and sets a new state-of-the-art on \dataset{POIC}, \dataset{POT-210}, and \dataset{POT-280}.
On \dataset{POT-210} it outperforms all other published methods by a large margin.
Inaccuracy of the \dataset{POT-210} ground truth accounted for half of the \method{WOFT} errors.
We publish the \method{WOFT} code, trained model and an improved GT annotation of a \dataset{POT-210} subset\footnote{\url{https://cmp.felk.cvut.cz/~serycjon/WOFT}}. \\
{\bf Acknowledgements.}
This work was supported by Toyota Motor Europe,
by CTU student grant SGS20/171/OHK3/3T/13, and
by the Research Center for Informatics project CZ.02.1.01/0.0/0.0/16\_019/0000765 funded by OP VVV.
{\small \bibliographystyle{ieee_fullname}
  \bibliography{sjo-bib} }

\end{document}